\documentclass[10pt,journal,compsoc]{IEEEtran}

\usepackage{times}
\usepackage{epsfig}
\usepackage{graphicx}

\usepackage{amsfonts}       
\usepackage{amsmath}
\usepackage{amssymb}
\usepackage{array}
\usepackage{booktabs}       
\usepackage{url}            
\usepackage{soul}
\usepackage{microtype}      
\usepackage{nicefrac}       

\usepackage{threeparttable}
\usepackage{extarrows}
\usepackage[table,xcdraw]{xcolor}
\usepackage{makecell}
\usepackage{algorithm, algpseudocode}

\usepackage{xcolor}
\usepackage{makeidx}
\usepackage{microtype}
\usepackage{url}
\usepackage{amsmath}
\usepackage{amssymb}
\usepackage{latexsym}
\usepackage{subfigure}
\usepackage{graphicx}
\usepackage{multirow}
\usepackage{color}
\usepackage{hyperref}
\usepackage{marginnote}
\usepackage{lipsum}

\usepackage{bm}
\usepackage{amsfonts,amssymb,amsmath,amsthm}

\usepackage{scalerel}
\usepackage{tikz}
\usetikzlibrary{svg.path}

\definecolor{orcidlogocol}{HTML}{A6CE39}
\tikzset{
  orcidlogo/.pic={
    \fill[orcidlogocol] svg{M256,128c0,70.7-57.3,128-128,128C57.3,256,0,198.7,0,128C0,57.3,57.3,0,128,0C198.7,0,256,57.3,256,128z};
    \fill[white] svg{M86.3,186.2H70.9V79.1h15.4v48.4V186.2z}
                 svg{M108.9,79.1h41.6c39.6,0,57,28.3,57,53.6c0,27.5-21.5,53.6-56.8,53.6h-41.8V79.1z M124.3,172.4h24.5c34.9,0,42.9-26.5,42.9-39.7c0-21.5-13.7-39.7-43.7-39.7h-23.7V172.4z}
                 svg{M88.7,56.8c0,5.5-4.5,10.1-10.1,10.1c-5.6,0-10.1-4.6-10.1-10.1c0-5.6,4.5-10.1,10.1-10.1C84.2,46.7,88.7,51.3,88.7,56.8z};
  }
}

\newcommand\orcidicon[1]{\href{https://orcid.org/#1}{\mbox{\scalerel*{
\begin{tikzpicture}[yscale=-1,transform shape]
\pic{orcidlogo};
\end{tikzpicture}
}{|}}}}

\newcommand{\x}{\bm{x}}

\newcommand{\Y}{\mathcal{Y}}

\def\ie{$i.e.$}
\def\eg{$e.g.$}
\def\etc{$etc$}

\long\def\comment#1{}

\def\red#1{\textcolor{red}{#1}}
\definecolor{blue}{rgb}{0,0,0.8}
\definecolor{green}{rgb}{0,0.4,0}
\definecolor{black}{rgb}{0,0,0}
\definecolor{mark}{rgb}{0,0,0}
\definecolor{mark2}{rgb}{0,0,0}
\definecolor{mark3}{rgb}{0,0,0}
\definecolor{rebuttal3}{rgb}{0,0,0}

\usepackage{times}
\usepackage{nicefrac}
\usepackage{graphics}
\usepackage{epsfig}
\usepackage{booktabs}
\usepackage{url}
\usepackage{psfrag}

\usepackage{amsfonts}
\usepackage{amsmath, amssymb}

\usepackage{epstopdf}
\usepackage{multirow}

\begin{document}
%
\title{Generalizable Black-Box Adversarial Attack with Meta Learning}
%
%
%
%

\author{Fei Yin$^{*}$\orcidicon{0000-0002-5146-7685},
        Yong Zhang$^{*}$\orcidicon{0000-0003-0066-3448},
        Baoyuan Wu$^{* \dagger}$\orcidicon{0000-0003-2183-5990},~\IEEEmembership{Member,~IEEE,}
        Yan Feng, 
        Jingyi Zhang,\\ 
        Yanbo Fan\orcidicon{0000-0002-8530-485X},
        Yujiu Yang$^{\dagger}$\orcidicon{0000-0002-6427-1024},~\IEEEmembership{Member,~IEEE} 
\IEEEcompsocitemizethanks{
\IEEEcompsocthanksitem 
Fei Yin, Yan Feng, and Yujiu Yang are with Tsinghua Shenzhen International Graduate School, Tsinghua University, Beijing 100190, China. E-mail: \{yinf20, y-feng18\}@mails.tsinghua.edu.cn, yang.yujiu@sz.tsinghua.edu.cn.
\IEEEcompsocthanksitem 
Baoyuan Wu is with the School of Data Science, Shenzhen Research Institute of Big Data, Chinese University of Hong Kong, Shenzhen 518172, China. E-mail: wubaoyuan@cuhk.edu.cn.
\IEEEcompsocthanksitem 
Yong Zhang and Yanbo Fan are with Tencent AI Lab, Shenzhen, Guangdong 518057, China. E-mail: \{zhangyong201303, fanyanbo0124\}@gmail.com.
\IEEEcompsocthanksitem
Jingyi Zhang is with the Center for Future Media and the School of Computer Science and Engineering, University of Electronic Science and Technology of China, Chengdu 610056, China. E-mail: jingyi.zhang1995@gmail.com.
\IEEEcompsocthanksitem
Fei Yin, Yong Zhang and Baoyuan Wu are co-first authors. Baoyuan Wu and Yujiu Yang are corresponding authors.
}
\thanks{
This work has been accepted by T-PAMI 2022.
This work was supported in part by the National Natural Science Foundation of China under Grant U1903213 and in part by the Shenzhen Key Laboratory of Marine IntelliSense and Computation under Grant ZDSYS20200811142605016. The work of Baoyuan Wu is supported by the Natural Science Foundation of China under Grant 62076213, Shenzhen Science and Technology Program under Grants RCYX20210609103057050 and ZDSYS20211021111415025, and in part by the University Development fund of the Chinese University of Hong Kong, Shenzhen under Grant 01001810, and CCF-Tencent Open Fund. \\ 
}
}

%
%

\markboth{Journal of \LaTeX\ Class Files,~Vol.~, No.~, }%
{Shell \MakeLowercase{\textit{et al.}}: Bare Demo of IEEEtran.cls for Computer Society Journals}
%




\IEEEtitleabstractindextext{

\begin{abstract}

In the scenario of black-box adversarial attack, the target model’s parameters are unknown, and the attacker aims to find a successful adversarial perturbation based on query feedback under a query budget. Due to the limited feedback information, existing query-based black-box attack methods often require many queries for attacking each benign example. To reduce query cost, we propose to utilize the feedback information across historical attacks, dubbed example-level adversarial transferability. Specifically, by treating the attack on each benign example as one task, we develop a meta-learning framework by training a meta generator to produce perturbations conditioned on benign examples. When attacking a new benign example, the meta generator can be quickly fine-tuned based on the feedback information of the new task as well as a few historical attacks to produce effective perturbations. Moreover, since the meta-train procedure consumes many queries to learn a generalizable generator, we utilize model-level adversarial transferability to train the meta generator on a white-box surrogate model, then transfer it to help the attack against the target model. The proposed framework with the two types of adversarial transferability can be naturally combined with any off-the-shelf query-based attack methods to boost their performance, which is verified by extensive experiments.
The source code is available at \url{https://github.com/SCLBD/MCG-Blackbox}. 
\end{abstract}

}

\maketitle

\IEEEdisplaynontitleabstractindextext

\IEEEpeerreviewmaketitle

\begin{figure*}[t]
\begin{center}
\centerline{\includegraphics[width=1\linewidth]{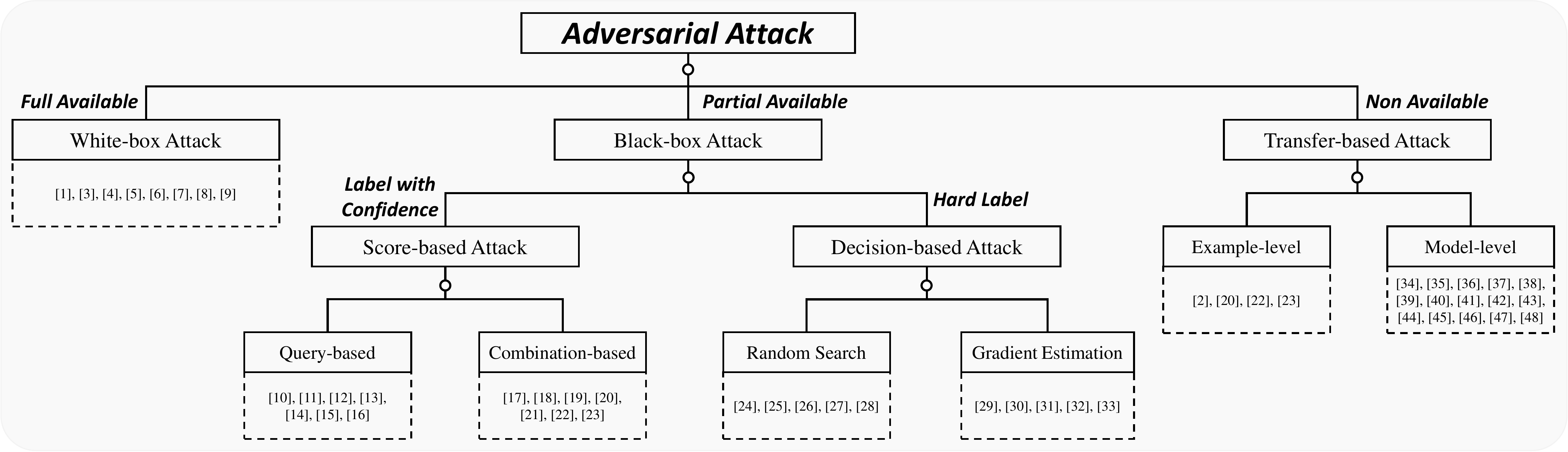}}
\vspace{-1mm}
\caption{Taxonomy structure of existing black-box adversarial attack methods.}
\label{fig:related_work}
\end{center}
\vspace{-4mm}
\end{figure*}

\IEEEraisesectionheading{\section{Introduction}
\label{sec:introduction}}

\IEEEPARstart{D}{eep} neural networks (DNNs) have been shown to be vulnerable to adversarial examples~\cite{GoodfellowSS14}, where stealthy and malicious perturbations are added onto benign examples to fool the DNN model. 
According to the accessible information about the attacked DNN model (\ie, the target model), existing adversarial attacks can be generally categorized into two scenarios. 
The first is \textit{white-box attack}, which assumes that the attacker knows the parameters of the target model, such that the adversarial perturbation can be easily generated based on the gradient of the target model. 
The second is \textit{black-box attack}, where the attacker does not know the parameters of the target model, while only the query feedback is accessible. 
Compared to the white-box scenario, the black-box scenario is more practical and more challenging.
\comment{, and \red{has not been studied as extensively}} 
Thus, this work focuses on the black-box scenario, especially on the \textit{score-based} black-box attack, where the feedback is a continuous score (\ie, the posterior probability in classification problem).

The general procedure of attacking one benign example in the score-based black-box attack scenario can be described as follows: 
given a query budget (\ie, the allowed query number) and an allowed perturbation region in the vicinity of the attacked benign example, with the starting solution at the benign example, the attacker keeps searching for a successful perturbation that satisfies the attack goal for a black-box target model; if such a successful perturbation is found within the allowed perturbation region under the query budget, then the attack is successful and stopped. Otherwise, the attack failed.

As image statistics differ in benign images, attacking a benign image can be viewed as an individual task where the feedback information from the target model serves as the supervision to guide the perturbation generation. 
This perspective inspires us to utilize the information across different tasks to learn generic prior knowledge that can be transferred to boost the performance of each individual task, as did in meta-learning. 
The intuition is that, when attacking more benign examples, the attacker is supposed to be more experienced in how to generate perturbation conditioned on a given benign image and also know the target model better. Consequently, compared to the fresh attacker, one experienced attacker is expected to find a successful perturbation with fewer queries when attacking a new benign example. 
The underlying rationale is that adversarial perturbations around different benign examples may have some similar properties, and we call it \textbf{example-level adversarial transferability}. 
One typical example is universal adversarial perturbations~\cite{moosavi2017universal}, where one perturbation may fool multiple benign examples simultaneously. 
However, to the best of our knowledge, 
example-level adversarial transferability has not been proposed explicitly to boost the attack performance, especially in the scenario of black-box attack.
To capture the above intuition, here we propose to utilize meta-learning to learn a meta generator across different attacking tasks (\ie, attacking different benign examples), dubbed \textit{Meta Conditional Generator (MCG)}, which encodes the prior into the parameters of the network and can produce adversarial perturbations based on the benign example accordingly. 
When attacking a new benign example, the meta generator can be quickly fine-tuned based on the information of the benign example as well as the feedback information of a few historical attacks to produce effective perturbations that are specific to the new benign example.

However, directly training the meta generator with the perturbations of successful attacks in the meta-train process may still require many queries to the target model. \comment{to learn a meta generator with good generalization to new tasks.} To further reduce the number of queries, we resort to the widely used \textbf{model-level adversarial transferability}, which assumes that some shared terms can be transferred from white-box surrogate models to the target model to help the black-box attack. 
Specifically, we propose to conduct meta training based on white-box surrogate models, then the learned generator is transferred to help the attack against the target model, which serves as the \textit{meta-test} process. 
Besides, to mitigate the difference between the surrogate and target models, we also fine-tune the surrogate model during the meta-test process by minimizing the feedback difference between the surrogate and target models for the same query, which encourages the surrogate model to mimic the behaviour of the target model and makes the generated perturbation more specific to the target model. The updated surrogate model is then exploited to fine-tune the meta generator for adaptation.
In short, inspired by the perspective of meta-learning, we propose a general meta-learning framework for black-box adversarial attacks, by utilizing two levels of adversarial transferability, including example-level and model-level.

One prominent advantage of the proposed framework is that it can be naturally combined with any off-the-shelf black-box attack methods to boost their original performance.
For example, for sampling-based methods, the meta generator can serve as the sampling distribution; for random-search-based methods that gradually adjust the perturbation, the meta generator can provide a suitably initialized perturbation. 
Extensive experiments on benchmark datasets and against several state-of-the-art attack methods verify the superiority of the proposed attack framework.

The main contributions of this work are three-fold. 
\textbf{1)} We propose to treat the black-box attack to each benign example as an individual task, which inspires us to utilize the information across different tasks to boost the attack performance of each task.
\textbf{2)} We develop a general meta-learning framework for the black-box attack scenario by utilizing both the example- and model-levels of adversarial transferability. 
\textbf{3)} Extensive experiments demonstrate that the proposed framework can be naturally combined with any existing query-based black-box attack methods and significantly boost their original performance.

\section{Related Work}
\label{sec: related work}

As shown in Fig.~\ref{fig:related_work}, we partition existing adversarial attack methods into three categories at the first levels, including \textbf{white-box methods} which utilize the full information (\ie, parameters) of the attacked model to generate perturbations, \textbf{black-box methods} which utilize the query feedback returned by the attacked model, and \textbf{transfer-based methods} which don't utilize any information of the attacked model (\ie, target model). 
Their detailed reviews are presented in the following sub-sections.

\subsection{White-box Adversarial Attack}

The white-box attack problem is generally formulated as follows:
\begin{flalign}
\max_{\bm{\delta} \in \mathbb{B}_{\epsilon}(\x)} \mathcal{L}(f(\x + \bm{\delta}), t), 
\label{eq: whitebox attack}
\end{flalign}
where $\bm{\delta}$ represents the adversarial perturbation, $\mathbb{B}_{\epsilon}(\x) = \{ \x' - \x | ~ \| \x' - \x \|_{p} \leq \epsilon \}$ denotes a neighboring region around $\x$, and the attacker-specified scalar $\epsilon > 0$ represents the upper bound of allowed perturbations.
The loss function $\mathcal{L}(f(\x + \bm{\delta}), t)$ could be specified as the cross entropy loss in untargeted attack where $t$ indicates the ground-truth label of $\x$, while as the negative cross entropy in targeted attack where $t$ indicates a target label that is different from the ground-truth label of $\x$. 
Many gradient-based optimization methods have been utilized to solve the above problem, including the fast gradient sign method (FSGM)~\cite{GoodfellowSS14}, iterative fast gradient sign method (I-FGSM)~\cite{kurakin2018adversarial}, C\&W method~\cite{carlini2017cwattack}, functional adversarial attack~\cite{laidlaw2019functional}, adversarial camouflage~\cite{duan2020adversarial}, \etc.
Moreover, several works~\cite{feng2021meta_pyhsical_attack,athalye2018synthesizing,jan2019connecting} focus on the adversarial attack in the physical world, where many types of environmental distortions may weaken the effectiveness of the generated adversarial perturbations.

\subsection{Black-box Adversarial Attack}
The formulation of Problem~\eqref{eq: whitebox attack} is also applicable to the black-box attack.
However, the parameters of the target model $f(\cdot)$ is unknown, while only the objective value $f(\bm{x} + \bm{\delta})$ for each query $\bm{x} + \bm{\delta}$ is provided. 
Consequently, the gradient-based optimization methods cannot be directly utilized.  
According to the type of query feedback $f(\bm{x} + \bm{\delta})$, black-box attacks can be further partitioned to two sub-categories, including score-based and decision-based attacks.

\subsubsection{Score-based Black-box Adversarial Attack}

In the scenario of score-based attack, the feedback $f(\x + \bm{\delta})$ is continuous, such as the posterior probability in the image classification task. 
Existing methods for solving this problem can be generally partitioned into three categories, including query-based and combination-based methods. 
\textbf{1) Query-based methods} iteratively adjust the perturbation only based on queries to the target model. Many black-box optimization approaches have been utilized, mainly including random search (\eg, \cite{GuoGYWW19}, Square, SignHunter \cite{Al-DujailiO20}), evolution strategies (\eg, NES \cite{IlyasEAL18}, Bandit \cite{IlyasEM19}), and gradient-estimation approaches (\eg, ZOO \cite{chen2017zoo}, AutoZoom \cite{tu2019autozoom}, ZO-signSGD \cite{liu2018signsgd}). 
Compared to transfer-based methods, query-based methods often achieve a higher attack success rate, but with the cost of many queries to the target model. 
\textbf{2) Combination-based methods} aim to achieve a high attack access rate and high query efficiency simultaneously, by taking advantage of both queries to the target model and the transferred item from the surrogate model. 
Existing methods proposed different kinds of transferring strategies, such as transferring the perturbation magnitude (\eg, Square \cite{ACFH2020square}), perturbation gradient (\eg, \cite{ChengDPSZ19}, \cite{GuoYZ19} and Meta attack \cite{DuZZYF20}), perturbation distribution (\eg, $\mathcal{N}$ATTACK \cite{LiLWZG19}, AdvFlow \cite{advflow}), the projection to a low-dimensional space (\eg, TREMBA \cite{Huang020}). 
Combination-based methods have shown superior attack performance to 
the other two categories, and our method also belongs to this category. However, most existing methods (only except Meta attack \cite{DuZZYF20}) only utilized the model-level adversarial transferability, while our method captures both example-level and model-level transferability to improve the query efficiency further.

\subsubsection{Decision-based Black-box Adversarial Attack}

In the scenario of decision-based attack, the feedback $f(\x + \bm{\delta})$ is discrete, such as the class label in the image classification task. 
Existing methods for solving this problem can be generally partitioned into random search based and gradient-estimation-based methods. 
\textbf{Random search based methods} aim to find the best perturbation around the invisible decision boundary, such as sampling from a normal distribution in Boundary method \cite{brendel2018decision} or from a learnable Gaussian distribution in Evolutionary method \cite{dong2019efficient}, searching along the estimated normal direction of the decision boundary in GeoDA \cite{rahmati2020geoda}, or searching on the surface of the allowed perturbation region in the vicinity of the benign example in SFA \cite{chen2020boosting} and Rays \cite{chen2020rays}. 
\textbf{Gradient-estimation-based methods} propose different estimation approaches of gradient, such as utilizing the neighboring points around the current solution in NES \cite{IlyasEAL18} and qFool \cite{LiuMF19}, Monte Carlo estimation in HopSkipJumpAttack \cite{chen2020hopskipjumpattack},  or estimating the gradient in a low-dimensional subspace for acceleration in QEBA \cite{abs-2005-14137}. 
OPT \cite{ChengLCZYH19} and Sign-OPT \cite{ChengSCC0H20} were developed based on a continuous formulation that alternatively optimizes the magnitude and direction of perturbation, such that any gradient-estimation approaches can be utilized.

\subsection{Transfer-based Adversarial Attack}
We list the transfer-based methods as another category, as each transfer-based method could be applied for the white-box attack or the black-box attack. For example, (MI-FGSM)~\cite{MIFGSM} and Nesterov iterative fast gradient sign method  (NI-FGSM)~\cite{lin2019nesterov} can be used as white-box attacks, since they are extensions of the classic white-box attacks FSGM~\cite{GoodfellowSS14} and I-FGSM~\cite{kurakin2018adversarial}. However, as claimed in the manuscripts of \cite{MIFGSM} and \cite{lin2019nesterov}, their goals are to generate more transferable perturbations to improve the attack success rate in the black-box settings, and their experiments contain both white-box and black-box settings. 
Thus, if one method mainly aims to improve the adversarial transferability, we partition it to the transfer-based category, rather than white-box or black-box. 
According to the transfer's objective, we further present example-level and model-level transferability, respectively.

\subsubsection{Example-level adversarial transferability} 

Although without explicit illustration, some works have actually studied the example-level transferability. 
For example, universal adversarial perturbations~\cite{moosavi2017universal} revealed that it is possible to find a perturbation to fool multiple benign examples simultaneously, with respect to the same attacked model. 
It reveals that different benign examples may have some common fragile directions, following which adversarial perturbations can be found easily. 
Another example is generation-based adversarial attacks~\cite{Huang020, advflow}, where a generative model is trained to directly generate an adversarial perturbation or adversarial example for each benign example. Its default assumption is that adversarial perturbations around different benign examples may follow the same distribution or mapping. 
However, the example-level adversarial transferability has been rarely utilized to boost the attack performance, especially in the scenario of black-box attack. The only attempt we have found is called Meta attack~\cite{DuZZYF20}, where a meta attacker is trained to generate gradient, which is then used as the update direction in gradient-estimation-based black-box attack methods. 
In contrast, our proposed meta-learning framework aims at capturing the distribution of adversarial perturbations, thus can be naturally combined with any kinds of query-based attack methods.

\subsubsection{Model-level adversarial transferability} 
\label{sec: subsec model-level transferability}
The model-level adversarial transferability has been observed and studied in many existing works~\cite{PapernotMG16, LiuCLS17, Huang020, inkawhich2020perturbing, Wu0X0M20, tramer2017space, yangcharacterizing}. 
Two important issues are mainly explored about the model-level transferability, including the intrinsic reason, and how to enhance the transferability across models, especially in the black-box scenario. 
\textbf{1) The intrinsic reason of the model-level transferability}.
Tram{\`e}r~\textit{et al.}~\cite{tramer2017space} found that different models share a large fraction of adversarial subspace, which consists of orthogonal basis vectors that are highly aligned with the gradient of the loss function. Consequently, perturbations within this shared subspace are likely to be transferable across models.  
A geometric perspective provided by \cite{LiuCLS17} showed that the transferability is partially due to the fact that decision boundaries of different models align well with each other. 
Demontis~\textit{et al.}~\cite{demontis2019adversarial} demonstrated that the model-level transferability is closely related to the intrinsic vulnerability of the target model, and the complexity of the surrogate model.
The theoretical analysis provided in \cite{yangcharacterizing} derives two lower bounds for transferability based on data distribution similarity and model gradient similarity, as well as the upper bound for transferability based on gradient orthogonality and smoothness.  
Wang~\textit{et al.}~\cite{wang2020unified} illustrated that the model-level transferability is negatively correlated with the interaction inside adversarial perturbations. 
\textbf{2) Enhancing the model-level transferability}.
Compared to FGSM~\cite{GoodfellowSS14}, its extensions, including momentum iterative fast gradient sign method (MI-FGSM)~\cite{MIFGSM} and Nesterov iterative fast gradient sign method (NI-FGSM)~\cite{lin2019nesterov}, showed better model-level transferability. 
In \cite{wang2021enhancing}, MI-FGSM and NI-FGSM were further extended by reducing the variance of the iterative update directions, such that the update direction is stabilized to escape from poor local optima, leading to the transferability improvement even when there are defenses for the target model. 
The ensemble attack method~\cite{LiuCLS17} generated adversarial perturbations based on multiple models to enhance the transferability for both untargeted and targeted attacks. 
Intermediate level attack (ILA)~\cite{huang2019enhancing} proposed to generate adversarial perturbations based on intermediate layers of surrogate models to avoid overfitting, such that the transferability to the target model can be enhanced. 
Feature distribution attack (FDA)~\cite{inkawhich2019transferable} focused on improving the transferability of a targeted attack by maximizing the activation of one intermediate layer between the benign example and the perturbed example. 
It is extended in \cite{inkawhich2020perturbing} from one intermediate layer to multiple intermediate layers, such that the transferability of targeted attack is further enhanced. 
Some works employed the meta-learning ideology to enhance transferability either.
MSM~\cite{qin2021meta_surrogate_model} obtained a Meta-Surrogate Model via optimizing a differentiable attacker. The Meta-surrogate model gained prior from one or a set of surrogate models and was able to generate adversarial examples with eximious transferability.
Meta Gradient~\cite{yuan2021meta_gradient} randomly sampled multiple models from a model zoo to compose different tasks and iteratively simulated a white-box attack and a black-box attack in each task, which narrowed the gap between the gradient directions in white-box and black-box attacks.

\section{The Proposed Approach}


\subsection{Problem Formulation}
\label{sec: subsec adversarial attack}

We denote the classification model as $f_{\boldsymbol{\theta}}: \mathcal{X} \rightarrow \mathcal{Y}$, where the model parameters $\boldsymbol{\theta}$ are unknown in the black-box attack setting, and $\mathcal{X}$ and $\mathcal{Y}$ are the input and output spaces, respectively. 
Given an input example $\x$, the index of its ground-truth label is denoted as $y$; 
$f_{\boldsymbol{\theta}}(\x, i) \in [0,1]$ indicates the posterior probability \textit{w.r.t.} the $i$-th class, and ${f_{\boldsymbol{\theta}}(\x, i)}$ denotes the corresponding logit. 
Our attack goal is to generate an adversarial perturbation $\boldsymbol{\delta}$ for the benign example $\x$ to fool the model $f_{\boldsymbol{\theta}}(\cdot)$, given that the model parameters $\boldsymbol{\theta}$ are unknown, dubbed 
\textit{score-based black-box adversarial attack}. 
It can be generally formulated as the following optimization problem: 
\begin{flalign}
    \min_{\bm{\delta} \in \mathbb{B}_{\epsilon}(\x)} 
    \mathcal{L}^{\text{adv}}(\bm{\delta}, \bm{x}, y) = 
    \begin{cases}
    \max(0, \triangle_{ut}
    ), & \text{if \it untargeted attack}
    \\
    \max(0, \triangle_t
    ), & \text{if \it targeted attack}
    \end{cases}
    \label{eq: black attack}
\end{flalign}
where $\triangle_{ut} = f_{\boldsymbol{\theta}}(\x + \bm{\delta}, y) - \max_{j\neq y} f_{\boldsymbol{\theta}}(\x + \bm{\delta}, j)$, and 
$\triangle_{t} = \max_{j\neq y} f_{\boldsymbol{\theta}}(\x + \bm{\delta}, j) - f_{\boldsymbol{\theta}}(\x + \bm{\delta}, t)$, with $t \in \Y$ being the target label. 
Note that $\mathcal{L}^{\text{adv}}$ is {\it non-negative}, and if $0$ is achieved, then the corresponding $\bm{\delta}$ is a successful adversarial perturbation.

\begin{figure}[t] 

\begin{center}
\centerline{\includegraphics[width=0.98\linewidth]{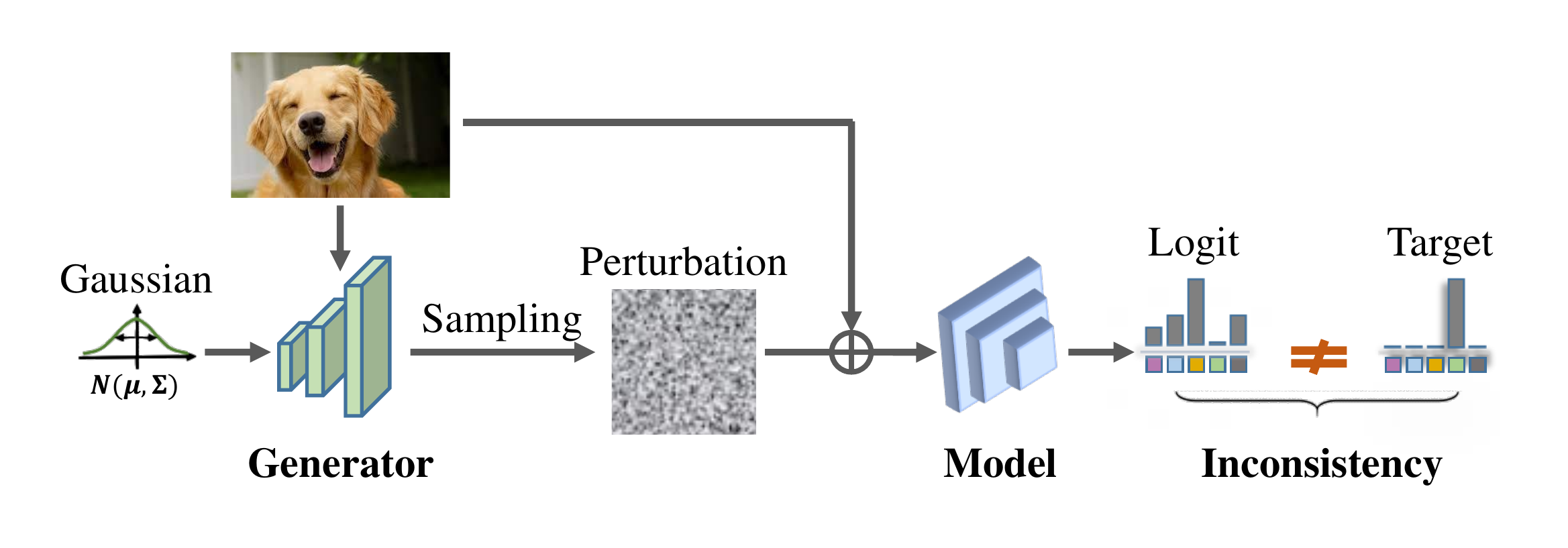}}
\vspace{-6mm}
\caption{Task definition in meta-learning. Given an image and a target model, the goal is to learn a generator that can generate an effective adversarial example to attack the target model.}
\label{fig: task}
\end{center}
\vspace{-4mm}
\end{figure}

\subsection{Meta Conditional Generator for Black-Box Attack} \label{sec:meta}
\vspace{1mm}
\textbf{Conditional Perturbation Generator.}
Unlike most previous combination-based methods that use a deterministic network to predict the initial perturbation for a benign image, our generator captures a conditional distribution of perturbation conditioned on the benign image, \textit{i.e.,} $
    {p}(\bm{\delta}|\bm{x};\boldsymbol{\varphi})$, where $ \bm{\delta}= G_{ \boldsymbol{\varphi}}(\bm{z}; \bm{x}) $.  $G$ is the generator with $\boldsymbol{\varphi}$ as its parameters. 
$\bm{\delta}$ is the perturbation and $\bm{z}$ is a random vector that follows a simple distribution, \textit{e.g.,} Gaussian distribution. 
In the conditional distribution, effective perturbations are supposed to have a high probability of being sampled. 
When attacking the target model, we can sample a perturbation $\bm{\delta} \sim {p}(\bm{\delta}|\bm{x};\boldsymbol{\varphi})$ and add it to the benign image to construct an adversarial example to fool the target model.

\begin{figure}[t] 

\begin{center}
\centerline{\includegraphics[width=1\linewidth]{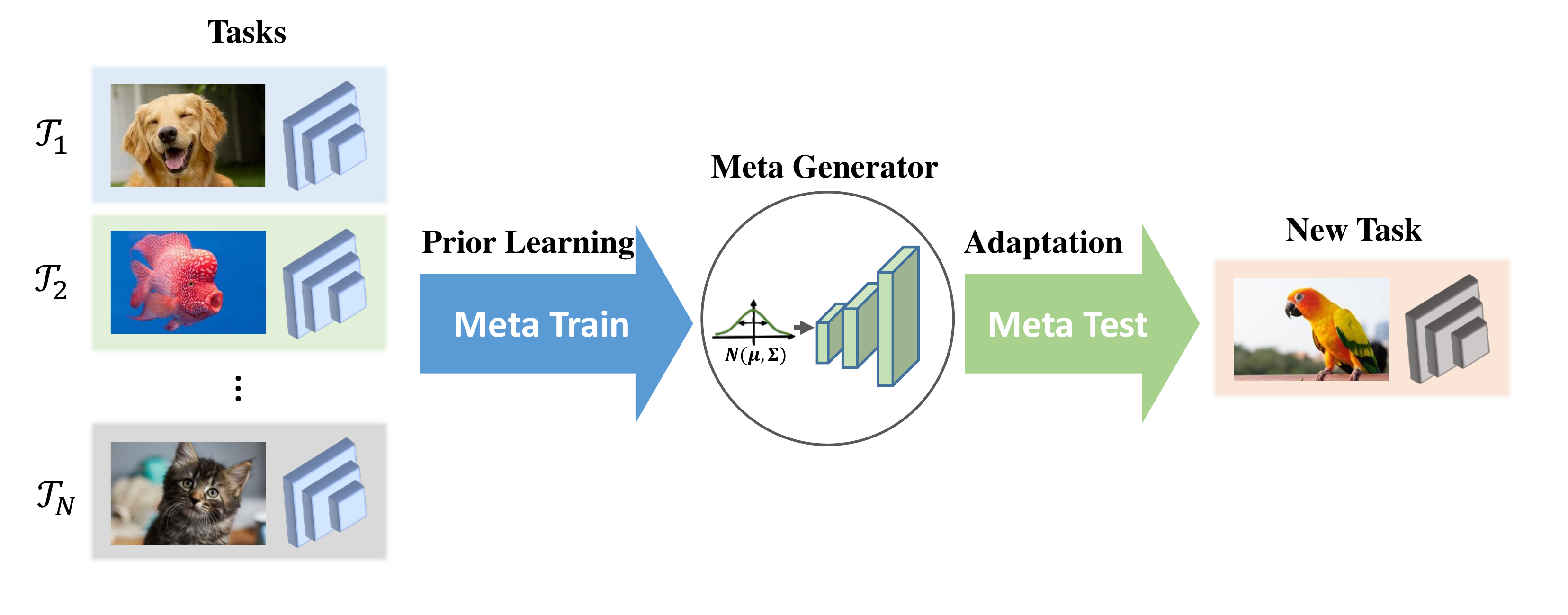}}
\vspace{-6mm}
\caption{Overview of meta-learning. 
During the meta-train phase, a meta generator can be obtained by training on a large set of tasks, which contains the generic prior of how to generate effective perturbations for different images to attack the target model. 
During the meta-test phase, the meta generator can be quickly adapted to new tasks with only a few steps of fine-tuning.}
\label{fig: meta}
\end{center}
\vspace{-4mm}
\end{figure}

\vspace{1mm}
\noindent \textbf{Modeling Conditional Distribution.}
The generator captures a conditional distribution of perturbation, which can be realized with using a simple distribution and a complex non-linear function that maps the simple distribution to a complex one with an image as the condition. 
In this work, we use a conditional generative flow (c-Glow)~\cite{cGlow} as the generator due to its superior property that the mapping between a random vector and the output perturbation is invertible.  
By using c-Glow, we have $\bm{\delta} = g_{\boldsymbol{\varphi}}(\bm{z}; \bm{x})$ and the inverse version $ \bm{z} = g^{-1}_{\boldsymbol{\varphi}}(\bm{\delta}; \bm{x})$. 
The random vector $\bm{z}$ follows a Gaussian distribution, \textit{i.e.,} $\bm{z} \sim \mathcal{N} (\bm{\mu}, \bm{\Sigma})$. 
Since c-Glow consists of a set of invertible functions/layers, the parameters $\boldsymbol{\varphi}$ can be decomposed into several individual parts, \textit{i.e.,} $ g = g_{\bm{x}, \varphi_1}  \circ \cdots  \circ g_{\bm{x}, \varphi_M} $. $M$ is the number of layers and $\varphi_i$ represents the parameters of the $i$-th layer.  
With the change of variables~\cite{Tabak2010Density}, we can write the conditional likelihood as
\begin{equation}
    \log {p}(\bm{\delta}|\bm{x};\boldsymbol{\varphi})=\log {p}(\bm{z})+\sum_{i=1}^{M} \log \left | \det(\frac{\partial g_{\varphi_i}^{-1}(\bm{r}_{i-1};\bm{x})}{\partial \bm{r}_{i-1}}) \right |,
    \label{eq:cglow_loss}
\end{equation}
where $\bm{r}_i = g_{\varphi_i}(\bm{r}_{i-1};\bm{x})$, $\bm{r}_0=\bm{x}$, and $\bm{r}_M = \bm{z}$.

\vspace{1mm}
\noindent \textbf{Perspective of Meta Learning.}
Attacking on one benign image can be viewed as an individual process where the generator can be optimized by sampling several perturbations from the conditional distribution and obtaining their corresponding feedback scores from the target model as supervision. 
However, when attacking a black-box model, the budget of query is always limited, \textit{i.e.,} we can only sample a few perturbations. 
Hence, the learning of the generator in each attacking process can be formulated as a few-shot learning problem, \textit{i.e.,} given $\{\bm{\delta}_i, f_{\bm{\theta}}(\bm{x}_i), y_i\}_{i=1}^{K}$, the goal is to learn the generator $G_{ \boldsymbol{\varphi}}(\bm{z}; \bm{x})$. $K$ is the number of shots.

As mentioned in Sec.~\ref{sec:introduction}, adversarial perturbations around different benign images may share certain common properties, \textit{i.e.,} example-level adversarial transferability. 
Inspired by the concept of ``learning to learn" in meta-learning, we can solve the few-shot learning problem from the perspective of meta-learning, and learn a meta generator to capture the common properties of perturbations by performing a large set of attacking tasks. 
The prior of how to generate a conditional distribution of perturbation around a benign image is learned across various and diverse tasks. 
Since the perturbation distributions of different benign images are generated through the same generator, the prior is implicitly encoded in the parameters of the generator.

\vspace{1mm}
\noindent \textbf{Task Definition.}
Task is the atom in meta-learning. 
In the scenario of adversarial attack, ``task" is defined as: ``\textit{given a benign image and a target model, the goal is to learn a conditional generative model from which a sampled perturbation could successfully fool the target model. }"
As shown in Fig.~\ref{fig: task}, given a benign image, we can sample a perturbation from the generative model. 
The addition of the benign image and the perturbation yields an adversarial example which is then fed into the target network for attack. 
The adversarial example is effective if its the predicted label is not consistent with the ground truth label (\textit{i.e.,} untargeted attack) or the predicted label of the adversarial example is the same as a specified label (\textit{i.e.,} targeted attack).  
Hence, the parameters of the generator are updated by maximizing the difference between the prediction and the ground truth or minimizing the difference between the prediction and the specified label.

Since the parameters and architecture of the target model are unknown, inspired by the transferability of adversarial example~\cite{PapernotMG16}, we use a surrogate model to replace the target model in the meta-train phase. 
As shown in Fig.~\ref{fig: meta}, we can achieve a meta generator by performing a large set of tasks, which captures the prior of generating adversarial perturbations and is adaptive to different tasks during the meta-test phase. 

\begin{figure}[t] 

\begin{center}
\centerline{\includegraphics[width=1\linewidth]{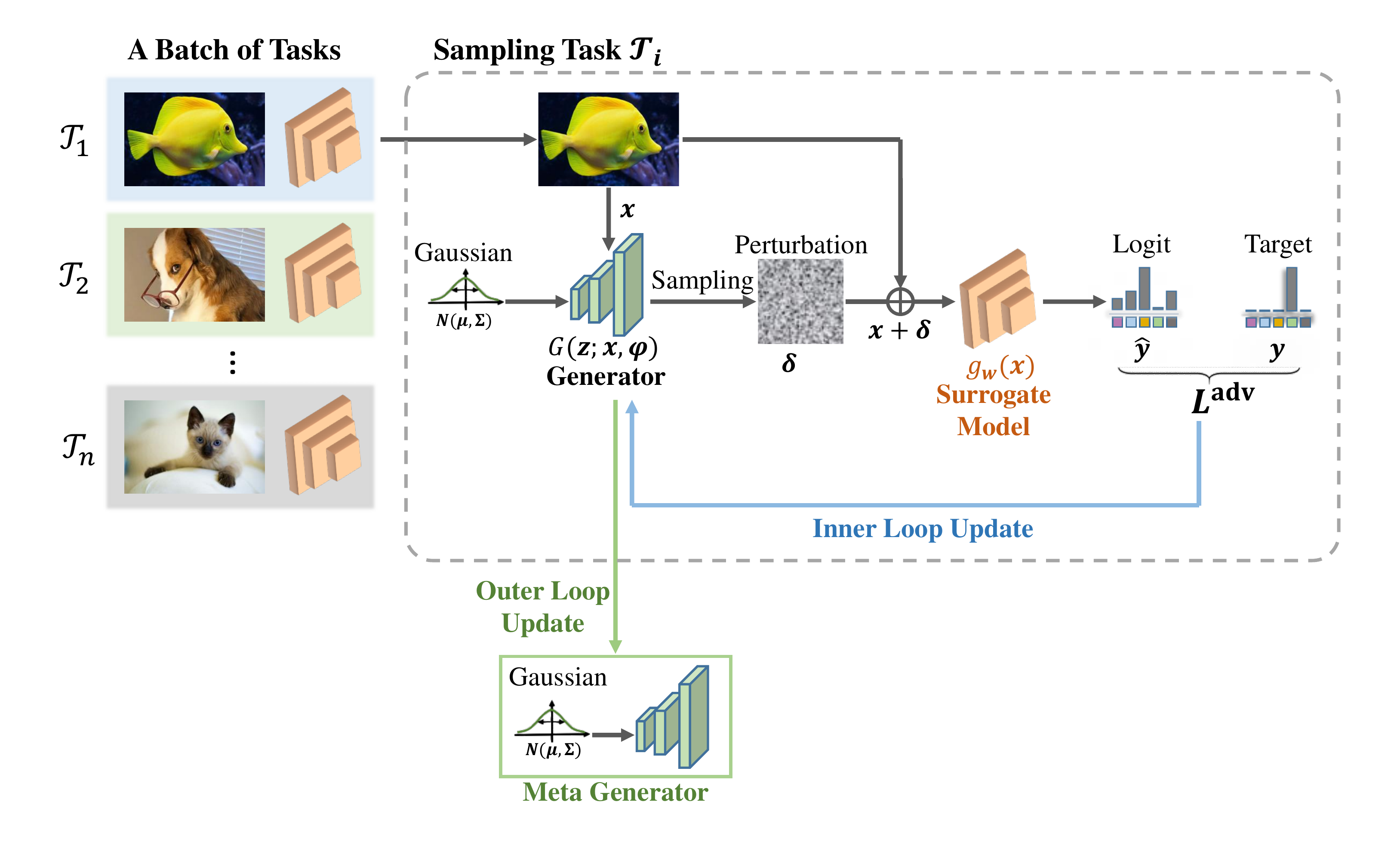}}
\vspace{-6mm}
\caption{The pipeline of meta training. 
The batch version of REPTILE is exploited for meta training.
We sample a batch of tasks and perform the inner loop update of task-specific parameters for each task. 
Then the task-specific parameters of all tasks in the batch are aggregated to do the outer loop update, \textit{i.e.,} updating the meta parameters. 
}
\label{fig: meta_train}
\end{center}
\vspace{-4mm}
\end{figure}

\begin{figure*}[t] 
\begin{center}
\centerline{\includegraphics[width=0.98\textwidth]{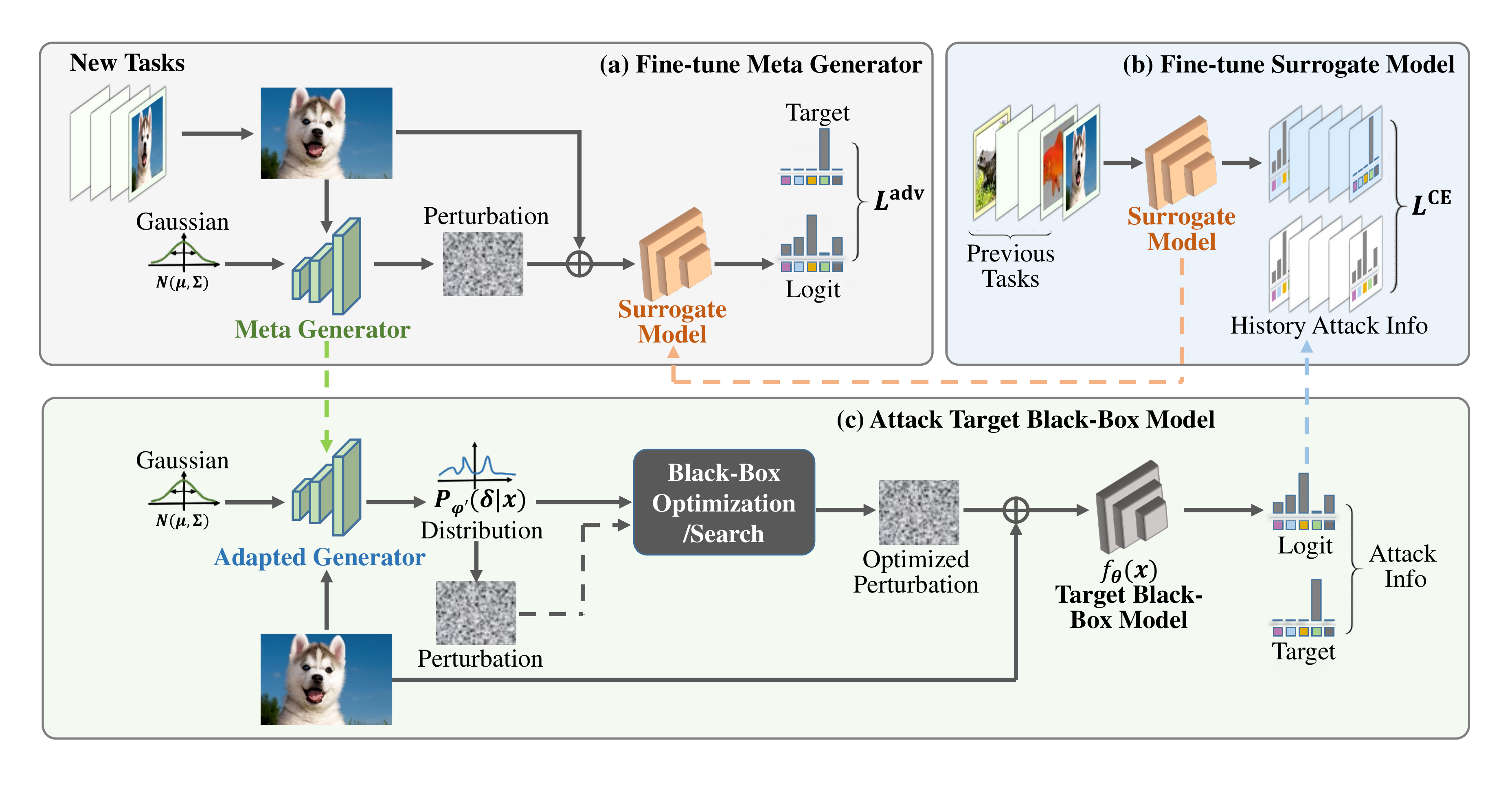}}
\vspace{-8mm}
\caption{
The pipeline of meta test. (a) Fine-tune the meta generator. Given the image of the 
new task and the updated surrogate model, the meta generator is fine-tuned by performing the inner optimization with $\mathcal{L}^{\text{adv}}$. 
(b) Fine-tune the surrogate model. In order to transfer the information of the target black-box model to the surrogate model, the historical information (\textit{i.e.,} logits of adversarial examples and the target labels) of previous tasks as well as the current task are used to make the surrogate model mimic the behaviour of the target model. 
(c) Attack the target black-box model. The adapted generator is combined with the off-the-shelf black-box attack methods. 
The generator can provide an initial distribution of perturbation or a sampled perturbation according to the given image. 
The initialization is leveraged by the attack methods as the starting state to get a refined perturbation. Then, the generated adversarial example is used to attack the target model. The logits of the attack are recorded to fine-tune the surrogate model in stage (b). 
}
\label{fig: metatest}
\end{center}
\vspace{-4mm}
\end{figure*}

\subsection{Meta Training with A Surrogate Model} \label{sec:train}
As described in Sec.~\ref{sec:meta}, in a task $\mathcal{T}$, given a benign image $\bm{x}$ and a target model $f_{ \boldsymbol{\theta}}(\bm{x},\cdot)$, the goal is to learn a conditional perturbation generator $G_{ \boldsymbol{\varphi}}(\bm{z}; \bm{x})$ that can generate an effective perturbation $ \bm{\delta}$ to fool the target model, \textit{e.g.,} $ f_{ \boldsymbol{\theta}}(\bm{x} + \bm{\delta},y) < \max_{j \neq y} f_{ \boldsymbol{\theta}}(\bm{x}+ \bm{\delta}, j)$ for untargeted attack and $ f_{ \boldsymbol{\theta}}(\bm{x}+ \bm{\delta},t) > \max_{j \neq t} f_{ \boldsymbol{\theta}}(\bm{x}+ \bm{\delta},j)$ for targeted attack, where $\bm{\delta} = G_{ \boldsymbol{\varphi}}(\bm{z}; \bm{x})$. 
The meta training process is illustrated in Fig.~\ref{fig: meta_train}. 
In this work, the target black-box model is the same for different tasks. 

In the scenario of black-box attack, we have no access to the parameters and the architecture of the target model. 
Hence, we have to make many queries for each benign image to generate successful perturbations and then use them to learn the meta generator, which is costly for query and hard to realize in real-world scenarios. 
Based on the model-level adversarial transferability, in the meta-train phase, the target model is replaced by a surrogate model $g_{\bm{w}}(\bm{x})$ of which the architecture and parameters are available. 
Therefore, the gradients can backpropagate through the surrogate model to update the generator. 
Since the meta-train phase does not involve any target model, once the meta training is over, the meta generator can be applied to any target model in the meta-test phase.

To evaluate the effectiveness of the generated perturbation, the adversarial loss function is similar to that in Eq.~(\ref{eq: black attack}). 
The difference is that $\tilde{\triangle}_{ut}$ and $\tilde{\triangle}_{t}$ are computed by using the surrogate model rather than the target model, \textit{i.e.,}
\begin{align}
    \tilde{\triangle}_{ut} &= g_{\bm{w}}(\x + \bm \delta, y) - \max_{j\neq y} g_{\bm{w}}(\x + \bm \delta, j), \nonumber \\
    \tilde{\triangle}_{t} &= \max_{j\neq y} g_{\bm{w}}(\x + \bm \delta, j) - g_{\bm{w}}(\x + \bm \delta, t),  \label{eq:loss}
\end{align}
where $t$ is the specified target class in targeted attack.

Given a set of tasks $\{\mathcal{T}_i\}_{i=1}^{N}$, we follow the batch version of REPTILE~\cite{Reptile} to perform meta-learning. 
We sample $n$ tasks to form a batch and update the task-specific parameters $k$ times using Adam~\cite{Adam} for each task. 
The objective of the inner loop optimization is $ \boldsymbol{\phi}_{\mathcal{T}_i} = \arg \min_{\boldsymbol{\phi}_{\mathcal{T}_i}} \mathcal{L}^{\text{adv}}_{\mathcal{T}_i}$. 
The optimization procedure can be represented as 
\begin{align}
    \boldsymbol{\phi}_{\mathcal{T}_i} = \text{Adam} (\mathcal{L}^{\text{adv}}_{\mathcal{T}_i}, \boldsymbol{\varphi}, k, \alpha), \label{eq:inner}
\end{align}
where $\boldsymbol{\phi}_{\mathcal{T}_i}$ is the final task-specific parameters of the generator after $k$ steps of performing Adam, starting from $\boldsymbol{\varphi}$.  
At each of the $k$ steps, a perturbation is sampled from the current conditional distribution. 
$\mathcal{L}^{\text{adv}}_{\mathcal{T}_i}$ is the adversarial loss of the $i$-th task. 
$\alpha$ is the learning rate of the inner loop.

Then, for the outer loop optimization, we can update the meta parameters of the generator with the resulted task-specific parameters in a mini-batch, \textit{i.e.,}
\begin{align}
\boldsymbol{\varphi} \leftarrow \boldsymbol{\varphi} + \beta \frac{1}{n} \sum_{i=1}^{n} \left ( \boldsymbol{\phi}_{\mathcal{T}_i} -\boldsymbol{\varphi}\right ),  \label{eq:outer}
\end{align}
where $\beta$ is the learning rate of the outer loop.

\subsection{Meta Test Using Historical Attack Experience} 
\label{sec:test}

The standard meta-test process is not applicable in black-box attack because the target model is unknown and the gradient cannot be backpropagated to fine-tune the meta generator through it. 
To solve this issue in the meta-test, we propose to transfer the information of the black-box target model to a surrogate model by using the feedback of previous attacks to fine-tune the surrogate model. 
The usage of the historical attack experience could make the surrogate imitate the behaviour of the target network, which provides more accurate and effective information to fine-tune the meta generator than directly using the surrogate model. 
The pipeline of meta-test is illustrated in Fig.~\ref{fig: metatest}. 
Given a new benign image, it consists of three steps to conduct the attack to the target model, \textit{i.e.,} fine-tuning the surrogate model for introducing the information of the target model, fine-tuning the meta generator for adaptation to the new benign image, and boosting off-the-shelf black-box attack methods. The attacking ability of the whole framework is gradually improved in the process of circulation. 

\vspace{1mm}
\noindent \textbf{Fine-tuning the Surrogate Model.} In Fig.~\ref{fig: metatest}(b), to use historical attack experience, the predicted logits of previous benign images, their adversarial examples, and the current benign image by the target model are collected to provide supervision for the fine-tuning of the surrogate model. 
The loss function is defined as 
\begin{align}
    \mathcal{L}^{\text{CE}} = ~& CE( g_{\bm{w}}(\bm{x}+\bm{\delta}), f_{\boldsymbol{\theta}} (\bm{x} + \bm{\delta})) \nonumber \\
     & + CE( g_{\bm{w}}(\bm{x}), f_{\boldsymbol{\theta}} (\bm{x}))
\end{align}
where $CE(\cdot)$ is the cross entropy loss function. 
The two terms represent the losses of the adversarial example and the benign image, respectively. 
For the benign image in the current task, only the second term is used. 
The optimization of the parameters of the surrogate model can be represented as 
\begin{align}
\bm{w}' = \text{Adam} ( \sum_i^m \mathcal{L}^{\text{CE}}_{\mathcal{T}_i}, \bm{w}, s, \lambda), 
\end{align}
where $\bm{w}'$ represents the parameters of the updated surrogate model. $\lambda$ is the learning rate and $m$ is the number of tasks in a mini-batch.  
$\mathcal{L}^{\text{CE}}_{\mathcal{T}_i}$ is the loss for the $i$-th task. $s$ is the number of update steps.

\vspace{1mm}
\noindent \textbf{Fine-tuning the Meta Generator.} 
Fig.~\ref{fig: metatest}(a) presents the fine-tuning of the meta generator with using the updated surrogate model $g_{\bm{w}'}(\bm{x})$ for the benign image of the new task.  
The fine-tuning procedure in the meta-test phase is similar to the inner optimization in the meta-train phase. 
We use the loss in Eq.~(\ref{eq:loss}) and Eq.~(\ref{eq:inner}) to update the parameters of the meta generator, \textit{i.e.,}
\begin{align}
  \boldsymbol{\phi}_\mathcal{T} = \text{Adam} (\mathcal{L}^{\text{adv}}_{\mathcal{T}}, \boldsymbol{\varphi}, k, \alpha), \label{eq: finetune}
\end{align}
where $\boldsymbol{\phi}_\mathcal{T} $ represents the adapted parameters for the new task. 
Different from Eq.~(\ref{eq:inner}), the updated surrogate model is used in  Eq.~(\ref{eq: finetune}) to yield out the loigts for the adversarial examples, \textit{i.e.,} $ g_{\bm{w}'}(\bm{x} + \bm{\delta})$. 
As the updated surrogate model contains the transferred information from the target model, it can provide more accurate and effective supervision to fine-tune the meta generator than the original surrogate model. 
Hence, the generated perturbation is more specific to the target model.

\vspace{1mm}
\noindent \textbf{Boosting Off-the-shelf Black-Box Attack Methods.} 
Our adapted generator can provide an initial distribution of perturbation or a perturbation conditioned on the given benign image, enabling it be combined with other off-the-shelf black-box attack methods to boost their original performance. 
Fig.~\ref{fig: metatest} (c) shows the process of attacking the target model. 
The initial distribution or perturbation is leveraged by a black-box attack method as the starting state. 
The further optimized perturbation is then added to the benign image to generate an adversarial example. 
The output logits from the target model are recorded as historical attack information, which is then used to fine-tune the surrogate model. 
When combined with sampling-based methods~\cite{IlyasEAL18, feng2020boosting}, the adapted generator served as a distribution. When combined with random-search-based methods~\cite{GuoGYWW19, Al-DujailiO20, ACFH2020square}, a perturbation can be a sample from the generator and serves as an initial state.

\section{Experiments}
\subsection{Experimental Settings} 
\label{sec:settings}
\noindent\textbf{Datasets.}
To demonstrate the effectiveness of the proposed method, we conduct comprehensive experiments on two commonly used benchmark databases, \textit{i.e.,} CIFAR-10~\cite{krizhevsky2009} and ImageNet~\cite{RussakovskyDSKS15}. 
Following the setting in \cite{feng2020boosting}, for the CIFAR-10 dataset, we randomly select $1,000$ images from the testing set for evaluation which cover all classes evenly. The images are resized to $32 \times 32$. 
For the ImageNet dataset, we first randomly select 10 classes from the $1,000$ classes and then use the $500$ images of each class from the validation set for evaluation.
The images are resized to $224 \times 224$. 
The target and surrogate models are trained on the training set of the corresponding dataset. 
On CIFAR-10, we use the full training set for meta-learning to learn the meta generator. On ImageNet, the training set of the 10 chosen classes are used for meta-learning.  

\vspace{1mm}
\noindent\textbf{Evaluation.}
We select $l_{\infty}$-based attacks and set the maximal distortion as $\epsilon = 0.031$ for CIFAR-10 and $\epsilon = 0.05$ for ImageNet with image pixel values re-scaled to $[0, 1]$. 
We set the maximal query budget to 10,000 times in all experiments.
If the attacker cannot successfully fool the target m odel within the query limit, we consider it a failure case.
Following the prior work~\cite{feng2020boosting}, we adopt the attack success rate (ASR), the mean query number (Mean), and the median query number (Median) of successful attacks to evaluate the attack performance.

\begin{table*}
	\centering
	\caption{Closed-set evaluation on the CIFAR-10 dataset.}
	\label{tab:cifar10_closedset}
	\vspace{-2mm}
	\scriptsize
	\begin{tabular*}{\hsize}{@{}@{\extracolsep{\fill}}lcccccccccccc@{}}
		\toprule
		Target model $\rightarrow$ &\multicolumn{3}{c}{ResNet-PreAct-110} & \multicolumn{3}{c}{DenseNet-121} & \multicolumn{3}{c}{VGG-19} & \multicolumn{3}{c}{PyramidNet-110} \\
		\cline{2-4}\cline{5-7}\cline{8-10}\cline{11-13}
		Attack Method $\downarrow$ & ASR & Mean & Median & ASR & Mean & Median & ASR & Mean & Median & ASR & Mean & Median \\
		\midrule
		\multicolumn{13}{c}{\textit{\textbf{Untargeted Attack}}} \\
		\midrule
		NES \cite{IlyasEAL18} & \bf100.0\% & 285.2 & 211.0 & 99.4\% & 430.8 & 274.0 &  99.1\% & 822.8 & 421.0 & \bf100.0\% & 287.9 & 190.0  \\
		MCG + NES & \bf100.0\% & \bf124.1 & \bf1.0 & \bf100.0\% & \bf133.2 & \bf1.0 &  \bf99.9\% & \bf371.7 & \bf1.0 & \bf100.0\% & \bf90.4 & \bf1.0 \\
		\midrule
		CG-Attack \cite{feng2020boosting} & \bf100.0\% & 230.0 & 81.0 & \bf100.0\% & 222.7 & 21.0 & \bf100.0\% & 386.4 & 101.0 & \bf100.0\% & 93.7 & \bf1.0 \\
		MCG + CG-Attack & \bf100.0\% & \bf112.3 & \bf1.0 & \bf100.0\% & \bf113.4 & \bf1.0 & \bf100.0\% & \bf262.7 & \bf1.0 & \bf100.0\% & \bf74.1 & \bf1.0 \\
		\midrule
		SimBA-DCT \cite{GuoGYWW19} & \bf100.0\% & 428.0 & 359.5 & \bf100.0\% & 449.9 & 352.0 & \bf99.7\% & 557.6 & 426.0 & \bf100.0\% & 341.1 & 273.5 \\
		MCG + SimBA-DCT & \bf100.0\% & \bf199.3 & \bf1.0 & \bf100.0\% & \bf127.0 & \bf1.0 & \bf99.7\% & \bf234.8 & \bf1.0 & \bf100.0\% & \bf114.6 & \bf1.0 \\
		\midrule
		Signhunter \cite{Al-DujailiO20} & \bf100.0\% & 167.0 & 83.0 & \bf100.0\% & 196.6 & 87.0 & \bf100.0\% & 238.3 & 99.0 & \bf100.0\% & 128.4 & 63.5 \\
		MCG + Signhunter & \bf100.0\% & \bf83.3 & \bf1.0 & \bf100.0\% & \bf61.3 & \bf1.0 & \bf100.0\% & \bf157.9 & \bf1.0 & \bf100.0\% & \bf26.9 & \bf1.0 \\
		\midrule
		Square \cite{ACFH2020square} & \bf100.0\% & 227.3 & 144.5 & \bf100.0\% & 260.3 & 159.0 & \bf100.0\% & 342.0 & 175.5 & \bf100.0\% & 165.5 & 100.5 \\
		MCG + Square & \bf100.0\% & \bf39.7 & \bf1.0 & \bf100.0\% & \bf47.6 & \bf1.0 & \bf100.0\% & \bf57.9 & \bf1.0 & \bf100.0\% & \bf29.6 & \bf1.0 \\
		\midrule
		MetaAttack~\cite{DuZZYF20} & \bf100.0\% & 642.6 & 632.0 & 99.8\% & 759.4 & 635.0 & \bf100.0\% & 686.4 & 633.0 & \bf100.0\% & 601.8 & 430.0 \\
		MCG + MetaAttack & \bf100.0\% & \bf204.6 & \bf1.0 & \bf100.0\% & \bf81.4 & \bf1.0 & \bf100.0\% & \bf188.6 & \bf1.0 & \bf100.0\% & \bf99.3 & \bf1.0 \\
		\midrule
		\multicolumn{13}{c}{\textit{\textbf{Targeted Attack}}} \\
		\midrule
		NES \cite{IlyasEAL18} & \bf100.0\% & 774.3 & 610.0 & 99.8\% & 1205.2 & 946.0 & 92.4\% & 2738.8 & 1954.0 & \bf100.0\% & 889.7 & 694.0 \\
		MCG + NES & \bf100.0\% & \bf591.8 & \bf443.0 & \bf100.0\% & \bf1009.2 & \bf800.0 & \bf98.5\% & \bf2688.6 & \bf1472.0 & \bf100.0\% & \bf657.8 & \bf506.0 \\
		\midrule
		CG-Attack \cite{feng2020boosting} & \bf100.0\% & 884.8 & 741.0 & \bf99.8\% & 859.7 & 681.0 & 96.8\% & 1476.9 & 901.0 & \bf100.0\% & 567.3 & 501.0 \\
		MCG + CG-Attack & \bf100.0\% & \bf430.8 & \bf181.0 & \bf99.8\% & \bf608.5 & \bf241.0 & \bf97.4\% & \bf944.1 & \bf381.0 & \bf100.0\% & \bf290.6 & \bf141.0 \\
		\midrule
		SimBA-DCT \cite{GuoGYWW19} & 99.6\% & 836.3 & 729.0 & 99.7\% & 944.8 & 842.0 & 98.6\% & 1170.3 & 986.0 & 99.8\% & 735.7 & 661.0 \\
		MCG + SimBA-DCT & \bf99.8\% & \bf664.2 & \bf623.5 & \bf99.8\% & \bf845.5 & \bf768.00 & \bf98.9\% & \bf983.3 & \bf861.0 & \bf99.9\% & \bf601.9 & \bf543.5 \\
		\midrule
		Signhunter \cite{Al-DujailiO20} & \bf100.0\% & 386.1 & 272.0 & \bf100.0\% & 465.1 & 323.0 & \bf100.0\% & 556.3 & 385.5 & \bf100.0\% & 320.9 & 232.0 \\
		MCG + Signhunter & \bf100.0\% & \bf267.3 & \bf124.5 & \bf100.0\% & \bf321.0 & \bf181.0 & \bf100.0\% & \bf399.1 & \bf210.0 & \bf100.0\% & \bf167.9 & \bf81.0 \\
		\midrule
		Square \cite{ACFH2020square} & 99.6\% & 504.7 & 369.0 & \bf100.0\% & 624.8 & 471.0 & \bf100.0\% & 827.2 & 593.5 & \bf100.0\% & 400.1 & 301.0 \\
		MCG + Square & \bf100.0\% & \bf92.3 & \bf20.0 & \bf100.0\% & \bf133.1 & \bf33.0 & \bf100.0\% & \bf141.7 & \bf22.0 & \bf100.0\% & \bf55.0 & \bf17.0 \\
		\midrule
		MetaAttack~\cite{DuZZYF20} & \bf100.0\% & 1174.7 & 899.0 & \bf100.0\% & 1294.5 & 1106.0 & 99.0\% & 1721.4 & 1106.0 & \bf100.0\% & 1065.9 & 890.0 \\
		MCG + MetaAttack & \bf100.0\% & \bf757.3 & \bf669.0 & \bf100.0\% & \bf992.9 & \bf887.0 & \bf100.0\% & \bf1180.7 & \bf882.0 & \bf100.0\% & \bf718.3 & \bf669.0 \\
		\bottomrule
	\end{tabular*}
\end{table*}

\vspace{1mm}
\noindent\textbf{Target and Surrogate Models.} 
On CIFAR-10, we consider four target models: ResNet-Preact-110~\cite{HeZRS16}, DenseNet-121~\cite{HuangLMW17}, VGG-19~\cite{SimonyanZ14a}, and PyramidNet-110~\cite{HanKK17}.
We follow the standard training process of image classification to obtain the checkpoints of these target models. 
The top-1 error rates of these four target models are $6.29\%$, $6.17\%$, $7.28\%$, and $7.51\%$ on the standard testing set, respectively. 
On ImageNet, we also evaluate our method on four target models: ResNet-18~\cite{HeZRS16}, VGG-16-BN~\cite{SimonyanZ14a}, WRN-50~\cite{ZagoruykoK16}, and InceptionV3~\cite{szegedy2016rethinking}. %
We use the official implementation of these methods and download their pre-trained checkpoints from torchvision. 
The top-1 error rates of these target models are $28.41\%$, $30.24\%$, $21.53\%$, and $22.71\%$ on the validation set of ImageNet, respectively. 
In all experiments, ResNet-18~\cite{HeZRS16} and  ResNet-50~\cite{HeZRS16} are used as the surrogate models on CIFAR-10 and ImageNet, respectively. The corresponding top-1 error rates of the surrogate models are $6.37\%$ for {ResNet-18} and $23.97\%$ for ResNet-50. 

To further verify the performance of our framework, we also conduct experiments of attacking black-box adversarial defense models on ImageNet, including JPEG-Compression-WideResNet-50~\cite{guo2018countering}, Small-Noise-Defense-WideResNet-50~\cite{SND}, FreeAdv-ResNet-50~\cite{freeadv}, and FastAdv-ResNet-50~\cite{fastadv}.

\vspace{1mm}
\noindent\textbf{Competing Methods.}
As our framework can provide a good initialization of perturbation, it can be treated as a plug-and-play component that can be combined with other black-box attack methods to boost their performance. 
In order to verify the versatility of our model, we combine our framework with 6 query-based black-box attack methods, including NES~\cite{IlyasEAL18}, CG-Attack~\cite{feng2020boosting}, SimBA~\cite{GuoGYWW19}, SignHunter~\cite{Al-DujailiO20},  Square~\cite{ACFH2020square}, and MetaAttack~\cite{DuZZYF20}. 
For search-based methods such as SignHunter, Square, SimBA, and MetaAttack, a sampled perturbation from the generator is the initialization. 
For sampling-based methods such as NES and CG-Attack, a distribution of perturbation represented by the generator is the initialization.

\begin{table*}

	\centering
	\caption{Closed-set evaluation on the ImageNet dataset.}
	\label{tab:imagenet_closedset}
	\vspace{-2mm}
	\scriptsize
	\begin{tabular*}{\hsize}{@{}@{\extracolsep{\fill}}lcccccccccccc@{}}
		\toprule
		Target Model $\rightarrow$ &\multicolumn{3}{c}{ResNet-18} & \multicolumn{3}{c}{VGG-16} & \multicolumn{3}{c}{WRN-50} &
		\multicolumn{3}{c}{Inception-V3} \\
		\cline{2-4}\cline{5-7}\cline{8-10}\cline{11-13}
		Attack Method $\downarrow$ & ASR & Mean & Median & ASR & Mean & Median & ASR & Mean & Median & ASR & Mean & Median \\
		\midrule
		\multicolumn{13}{c}{\textit{\textbf{Untargeted Attack}}} \\
		\midrule
		NES~\cite{IlyasEAL18} & 99.0\% & 2085.5 & 1597.0 & 98.2\% & 1554.0 & 1072.0 & 96.5\% & 2366.0 & 1681.0 & 95.8\% & 1111.6 & 526.0 \\
		MCG + NES & \bf99.7\% & \bf1057.5 & \bf1.0 & \bf99.1\% & \bf596.5 & \bf1.0 & \bf98.4\% & \bf1385.1 & \bf1.0 & \bf96.7\% & \bf889.9 & \bf390.5 \\
		\midrule
		CG-Attack~\cite{feng2020boosting} & 96.3\% & 272.0 & \bf21.0 & 96.5\% & 178.3 & 21.0 & 89.6\% & 323.0 & \bf21.0 & 94.1\% & \textbf{251.1} & \bf21.0 \\
		MCG + CG-Attack & \bf100.0\% & \bf95.4 & \bf21.0 & \bf99.7\% & \bf69.3 & \bf1.0 & \bf99.2\% & \bf247.3 & \bf21.0 & \bf97.0\% & 271.1 & \bf21.0 \\
		\midrule
		SimBA-DCT~\cite{GuoGYWW19} & \bf100.0\% & 520.6 & 381 & 98.3\% & 486.1 & 350.0 & \bf94.9\% & 810.6 & 593.0 & 81.9\% & 496.9 & 290.5 \\
		MCG + SimBA-DCT & 98.1\% & \bf68.7 & \bf1.0 & \bf98.8\% & \bf73.6 & \bf1.0 & 94.7\% & \bf159.6 & \bf1.0 & \bf95.0\% & \bf129.2 & \bf9.5 \\
		\midrule
		Signhunter~\cite{Al-DujailiO20} & \bf100.0\% & 60.2 & 23.0 & \bf100.0\% & 69.8 & 34.0 & \bf100.0\% & 116.0 & 38.5 & 99.4\% & 178.3 & 49.0 \\
		MCG + Signhunter & \bf100.0\% & \bf29.6 & \bf1.0 & \bf100.0\% & \bf19.0 & \bf1.0 & \bf100.0\% & \bf62.8 & \bf1.0 & \bf100.0\% & \bf91.8 & \bf8.0 \\
		\midrule
		Square~\cite{ACFH2020square} & \bf100.0\% & 72.8 & 26.0 & \bf100.0\% & 82.6 & 28.0 & \bf100.0\% & 136.1 & 68.0 & 99.4\% & 210.8 & 64.0 \\
		MCG + Square & \bf100.0\% & \bf31.7 & \bf1.0 & \bf100.0\% & \bf24.8 & \bf1.0 & \bf100.0\% & \bf59.9 & \bf1.0 & \bf100.0\% & \bf123.8 & \bf24.0 \\
		\midrule
		MetaAttack~\cite{DuZZYF20} & 92.2\% & 3302.6 & 2641.0 & 95.4\% & 3075.2 & 2213.0 & 87.4\% & 3824.2 & 3309.0 & 94.7\% & 2634.8 & 1772.0 \\
		MCG + MetaAttack & \bf94.1\% & \bf1692.3 & \bf1.0 & \bf96.8\% & \bf1025.9 & \bf1.0 & \bf92.5\% & \bf2076.8 & \bf1.0 & \bf95.3\% & \bf2107.1 & \bf1324.0 \\
		\midrule
		\multicolumn{13}{c}{\textbf{\textit{Targeted Attack}}} \\
		\midrule
		NES~\cite{IlyasEAL18} & 67.9\% & 5734.8 & 5860.0 & 79.4\% & 4944.1 & 4621.0 & 33.0\% & 6137.9 & 6259.0 & 63.3\% & 4921.1 & 4726.0 \\
		MCG + NES & \bf75.3\% & \bf5499.6 & \bf5294.0 & \bf81.0\% & \bf4720.9 & \bf4559.0 & \bf37.5\% & \bf5839.1 & \bf5882.0 & \bf66.2\% & \bf4687.7 & \bf4370.0 \\
		\midrule
		CG-Attack~\cite{feng2020boosting} & 96.9\% & 2553.3 & 1801.0 & \bf92.9\% & 2447.7 & 1731.0 & 77.8\% & 2976.5 & 2191.0 & 91.0\% & \bf2260.3 & 1481.0 \\
		MCG + CG-Attack & \bf98.0\% & \bf2501.8 & \bf1781.0 & 91.9\% & \bf2374.4 & \bf1721.0 & \bf79.8\% & \bf2960.2 & \bf2101.0 & \bf94.9\% & 2272.1 & \bf1441.0 \\
		\midrule
		SimBA-DCT~\cite{GuoGYWW19} & 56.0\% & 6927.2 & 7196.0 & 71.7\% & 6569.9 & 6507.0 & 41.0\% & 6795.4 & 7028.0 & 56.8\% & 6094.9 & 6107.0 \\
		MCG + SimBA-DCT & \bf66.8\% & \bf6460.8 & \bf6607.0 & \bf79.3\% & \bf6023.3 & \bf6038.0 & \bf60.6\% & \bf5744.8 & \bf5626.0 & \bf72.8\% & \bf5576.9 & \bf5564.0 \\
		\midrule
		Signhunter~\cite{Al-DujailiO20} & \bf100.0\% & 1332.7 & 922.0 & \bf100.0\% & 1115.8 & 734.0 & 99.4\% & 1786.8 & 1064.0 & \bf100.0\% & 1836.2 & 1063.0 \\
		MCG + Signhunter & \bf100.0\% & \bf1043.8 & \bf596.5 & \bf100.0\% & \bf788.8 & \bf446.0 & \bf99.7\% & \bf1428.4 & \bf782.0 & 99.7\% & \bf1486.7 & \bf865.0 \\
		\midrule
		Square~\cite{ACFH2020square} & \bf100.0\% & 1097.7 & 819.0 & \bf100.0\% & 1112.7 & 819.0 & 99.7\% & 1678.7 & 1242.0 & 99.0\% & 1789.0 & 1120.0 \\
		MCG + Square & \bf100.0\% & \bf797.6 & \bf518.5 & \bf100.0\% & \bf784.4 & \bf529.0 & \bf100.0\% & \bf1148.9 & \bf661.0 & \bf99.7\% & \bf1455.2 & \bf868.0 \\
		\midrule
		MetaAttack~\cite{DuZZYF20} & 33.8\% & 8202.6 & 8041.5 & 17.5\% & 8341.4 & 8806.0 & 10.3\% & 8408.4 & 8585.0 & \bf19.5\% & 7176.8 & 8141.5 \\
		MCG + MetaAttack & \bf45.6\% & \bf7315.1 & \bf8033.0 & \bf38.1\% & \bf6425.6 & \bf7721.0 & \bf20.6\% & \bf7029.9 & \bf7596.5 & \bf19.5\% & \bf6946.5 & \bf7706.0 \\
		\bottomrule
	\end{tabular*}
\end{table*}

\begin{figure*}[t] 
\begin{center}
\centerline{\includegraphics[width=1\linewidth]{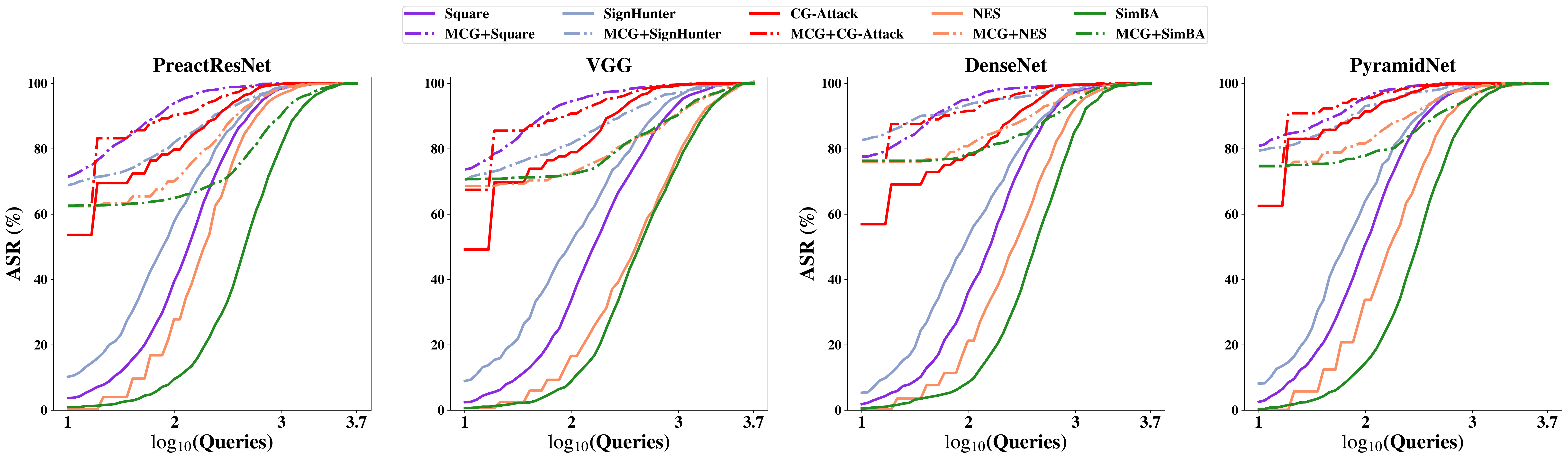}}
\vspace{-4mm}
\caption{Attack success rate (ASR\%) \textit{w.r.t.} the query number of untargeted attack on the CIFAR-10 dataset.}
\label{fig: cifar_query_curve}
\end{center}
\vspace{-8mm}
\end{figure*}

We also compare with several transfer-based attack methods to verify the transferability of the proposed method, including PGD~\cite{PGD}, MI~\cite{MIFGSM}, TIMI~\cite{TIMIFGSM}, and DI~\cite{DIFGSM}, under the transfer attack setting where only the information of the surrogate model is accessible. 
Moreover, we finally compare with several combination-based methods under the query attack setting, including AdvFlow~\cite{advflow} and TREMBA~\cite{Huang020}. 
They exploit both the queries to the target model and the transferred item from the surrogate model. 
To ensure the fairness of comparison, we retrain the combination-based methods with the same surrogate model as ours.
All the experiments are implemented with the source code provided by their authors under the same setting.

\vspace{1mm}
\noindent\textbf{Implementation Details.} \label{sec:exp_setting}
Following \cite{cGlow}, in all the experiments, we adopt the same architecture for the generator c-Glow with 3 blocks composed of 8 flow steps. 
Each block starts with a squeeze operation followed by 8 flow steps and ends with a split operation. 
To improve the efficiency, we adopt the discrete cosine transform (DCT) and inverse DCT for dimension reduction by downsampling the size of images in ImageNet to $\frac{1}{8} \times \frac{1}{8}$ lower frequency subspace before feeding them into the generator. 
On CIFAR-10, we use the original shape of images as they are in a small size. 

Before meta training, we pre-train the c-Glow to provide an initial state of modelling the distribution of perturbation. 
We use the surrogate model to generate a large set of perturbations through PGD attack with the perturbation strength of $\ell_{\inf} = 0.05$, the step size of $0.01$, and the number of iterations of $50$.  
The parameters are optimized by maximizing the log-likelihood, \textit{i.e.,}
$\max_{\boldsymbol{\varphi}} \log {p}(\bm{\delta}|\bm{x}; \boldsymbol{\varphi})$.

For the pre-training of the generator, the learning rate is set to $0.001$. The batch size is $m=16$. The generator is trained for 10 epochs. 
For meta training, we sample $16$ tasks every batch. 
The update stepsize of the inner optimization is set to $k=4$. 
The learning rates of the inner and outer loops are set to $\alpha=0.0003$ and $\beta=0.0006$, respectively. 
For meta-test, when fine-tuning the surrogate model, we freeze the parameters of all layers except the last three layers. 
The surrogate model is fine-tuned with both benign images and their adversarial examples. The learning rate of fine-tuning is set to $\lambda=0.0003$. 
The number of benign images is set to 4 in a batch. 
When fine-tuning meta generator, the process is similar to the inner loop optimization of the task-specific parameters.

\subsection{Experiments in Closed-set Attack Scenario}
In the closed-set attack scenario, the surrogate and target models are trained on the same training set, \textit{i.e.,} both the training images and the categories are the same. 
Experiments includes boosting the off-the-shelf black-box attack methods, attacking defended models, comparisons with transfer-based methods, and comparisons with combination-based methods.

\vspace{1mm}
\noindent\textbf{Performance on CIFAR-10.}
The specification of the surrogate model and the target models is presented in Sec.~\ref{sec:settings}. 
Table~\ref{tab:cifar10_closedset} illustrates the results of several off-the-shelf black-box attack methods and those of the combinations with our proposed MCG.

\begin{table*}
	\centering
	
	\caption{Untargeted Attack against adversarial defended models on the ImageNet dataset.}
	\label{tab:imagenet_defense}
	\vspace{-2mm}
	\scriptsize
	\begin{tabular*}{\hsize}{@{}@{\extracolsep{\fill}}lcccccccccccc@{}}
		\toprule
		Target Model $\rightarrow$ &\multicolumn{3}{c}{JPEG-Compress-WRN-50} & \multicolumn{3}{c}{SND-WRN-50} & \multicolumn{3}{c}{FastAdv-ResNet-50} & \multicolumn{3}{c}{FreeAdv-ResNet-50} \\
		\cline{2-4}\cline{5-7}\cline{8-10}\cline{11-13}
		Attack Method $\downarrow$ & ASR & Mean & Median & ASR & Mean & Median & ASR & Mean & Median & ASR & Mean & Median \\
		\midrule
		NES~\cite{IlyasEAL18} & 13.2\% & 5682.1 & 3243.0 & 76.6\% & 3879.8 & 3308.5 & \bf23.5\% & 7501.5 & 7986.5 & 15.7\% & 7188.5 & 6322.0 \\
		MCG + NES & \bf98.9\% & \bf1412.1 & \bf1.0 & \bf81.7\% & \bf1795.7 & \bf1.0 & \bf23.5\% & \bf2415.4 & \bf715.0 & \bf22.4\% & \bf2104.1 & \bf685.0 \\
		\midrule
		CG-Attack~\cite{feng2020boosting} & 39.0\% & 2227.5 & 1161.0 & \bf79.8\% & 540.5 & 81.0 & 58.5\% & 1789.7 & 871.0 & 61.3\% & 2374.0 & 1121.0 \\
		MCG + CG-Attack & \bf91.7\% & \bf180.9 & \bf1.0 & \bf79.8\% & \bf331.5 & \bf1.0 & \bf64.3\% & \bf1634.4 & \bf801.0 & \bf64.7\% & \bf2305.1 & \bf1081.0 \\
		\midrule
		SimBA-DCT~\cite{GuoGYWW19} & 14.3\% & 4688.3 & 4545.0 & 10.5\% & 608.3 & 22.0 & \bf45.5\% & 702.5 & 72.5 & \bf52.7\% & \bf792.5 & 245.0 \\
		MCG + SimBA-DCT & \bf67.1\% & \bf543.9 & \bf1.0 & \bf60.4\% & \bf498.3 & \bf1.0 & \bf45.5\% & \bf634.4 & \bf34.0 & \bf52.7\% & 821.9 & \bf77.0 \\
		\midrule
		Signhunter~\cite{Al-DujailiO20} & \bf100.0\% & 120.7 & 39.0 & \bf89.4\% & 133.4 & 31.0 & \bf89.4\% & 1101.5 & 37.5 & \bf88.2\% & \bf931.3 & 42.0 \\
		MCG + Signhunter & \bf100.0\% & \bf69.8 & \bf1.0 & 89.1\% & \bf65.6 & \bf1.0 & \bf89.4\% & \bf1083.5 & \bf36.5 & 86.5\% & 1058.3 & \bf40.0 \\
		\midrule
		Square~\cite{ACFH2020square} & \bf100.0\% & 140.9 & 71.0 & 88.9\% & 1487.8 & 154.0 & \bf92.4\% & 940.5 & 182.0 & \bf91.4\% & 871.3 & 170.0 \\
		MCG + Square & \bf100.0\% & \bf57.5 & \bf1.0 & \bf90.7\% & \bf307.4 & \bf1.0 & \bf92.4\% & \bf864.9 & \bf137.0 & \bf91.4\% & \bf724.9 & \bf123.5 \\
		\midrule
		MetaAttack~\cite{DuZZYF20} & 8.0\% & 3515.9 & 1557.0 & 11.7\% & 2449.1 & 1777.0 & 53.8\% & 4832.2 & 5063.5 & 55.3\% & 4614.9 & 4631.0 \\
		MCG + MetaAttack & \bf50.9\% & \bf117.2 & \bf1.0 & \bf53.1\% & \bf535.7 & \bf1.0 & \bf55.7\% & \bf4119.9 & \bf3969.5 & \bf56.3\% & \bf3580.2 & \bf3094.0 \\
		\bottomrule
	\end{tabular*}
\end{table*}

\begin{table*}[t] 
\vspace{-0.3em}
\centering

\caption{{Comparison with combination-based methods on the ImageNet dataset}}
\label{tab:imagenet_combination}
\vspace{-2mm}
\scriptsize
\begin{tabular*}{\hsize}{@{}@{\extracolsep{\fill}}lcccccccccccc@{}}
	\toprule
	Target Model $\rightarrow$ &\multicolumn{3}{c}{ResNet-18} & \multicolumn{3}{c}{VGG-16} & \multicolumn{3}{c}{WRN-50} &
	\multicolumn{3}{c}{Inception-V3} \\
	\cline{2-4}\cline{5-7}\cline{8-10}\cline{11-13}
	Attack Method $\downarrow$ & ASR & Mean & Median & ASR & Mean & Median & ASR & Mean & Median & ASR & Mean & Median \\
	\midrule
	\multicolumn{13}{c}{\textit{\textbf{Untargeted Attack}}} \\
	\midrule
    TREMBA~\cite{Huang020} & \bf100.0\% & 664.5 & 169.0 & 99.1\% & 160.2 & \bf1.0 & 99.0\% & 697.6 & 148.0 & 96.9\% & 1232.7 & 211.0 \\
    AdvFlow~\cite{advflow} & \bf100.0\% & 578.8 & 400.0 & \bf100.0\% & 693.3 & 420.0 & \bf100.0\% & 937.9 & 400.0 & \bf97.2\% & 716.3 & 200.0 \\
    MCG + CG-Attack & \textbf{100.0\%} & \textbf{95.4} & \textbf{21.0} & 99.7\% & \textbf{69.3} & \textbf{1.0} & 99.2\% & \textbf{247.3} & \textbf{21.0} & 97.0\% & \textbf{271.1} & \textbf{21.0} \\
    \midrule
    \multicolumn{13}{c}{\textit{\textbf{Targeted Attack}}} \\
    \midrule
    TREMBA~\cite{Huang020} & 81.4\% & 3197.2 & 1997.5 & 91.1\% & 2493.5 & 1759.0 & 60.2\% & 5140.5 & 3529.0 & 13.3\% & 9462.1 & 9433.0 \\
    AdvFlow~\cite{advflow} & 83.8\% & 4650.0 & 4400.0 & 81.6\% & 4407.8 & 4200.0 & 61.4\% & 5210.6 & 4980.0 & 81.8\% & 3734.1 &  3200.0 \\
    MCG + CG-Attack & \textbf{98.0\%} & \textbf{2501.8} & \textbf{1781.0} & \textbf{91.9\%} & \textbf{2374.4} & \textbf{1721.0} & \textbf{79.8\%} & \textbf{2960.2} & \textbf{2101.0} & \textbf{94.9\%} & \textbf{2272.1} & \textbf{1441.0} \\
\bottomrule
\end{tabular*}
\end{table*}

The MCG can boost the attack efficiency of all the black-box attack methods without ASR drop under both untargeted and targeted attacks.
Specifically, under the untargeted attack setting, the median query numbers of these methods are decreased to $1$s for all the target models by using the initialization from our MCG, which means that we can fool the target models with the initially generated perturbations for over $50\%$ images. Meanwhile, the ASRs are improved to nearly $100\%$ for almost all the cases. 
Besides, the mean query numbers are also improved significantly. 
For example, for the \emph{state-of-the-art} Square attack, our MCG can further improve its query efficiency in terms of the mean query number by a factor of 5 for all the four target models. 
We further plot the tendency curves of ASR \textit{w.r.t.} the query number in Fig.~\ref{fig: cifar_query_curve}. 
We can observe that the MCG can boost the attack performance under all values of query number for all the competing attack methods, especially for small query numbers.

Under the targeted attack setting, our proposed MCG can also boost the attack performance.
The targeted attack is generally harder to achieve than the untargeted attack, yet our MCG still obtains a satisfactory ASR attacking all the four target models.
The best attack performance is achieved by MCG+Square.
Compared to the original Square attack, the MCG+Square achieves the ASR of 100\% with a significantly fewer number of queries.
For example, the mean and median query numbers of MCG are over 4.6 times and 14.3 times less than those of the original Square attack.

\vspace{1mm}
\noindent \textbf{Performance on ImageNet.}
ImageNet is a much larger dataset than CIFAR-10.
The performance of both untargeted and targeted attacks on ImageNet is reported in Table~\ref{tab:imagenet_closedset}. 

We have a similar observation that the proposed MCG obtains consistent improvements when combined with different attack methods.
Under the untargeted attack setting, the improvement of ASR in several cases can be apparently observed.  
For example, when attacking WRN-50 and VGG-16 with CG-Attack, the ASRs are $89.6\%$ and $96.5\%$, respectively. 
Our MCG further improves CG-Attack by about $10\%$ for attacking WRN-50 and $ 3\%$ for attacking VGG-16.  
Besides, the efficiency improvements are also noticeable, especially for NES and SimBA. 
Under the targeted attack setting, the MCG brings in different degrees of improvements in almost all cases. The increase in ASR is more evident than the untargeted attack setting. 
Although when combined with SignHunter against InceptionV3, the ASR has been slightly reduced ($0.3\%$ lower) while the attack cost is saved near $20\%$.

All these results demonstrate that our MCG can provide an effective initial perturbation or a distribution of perturbation for various off-the-shelf black-box attack methods to boost their performance. 
Please note that the meta training procedure of the meta generator does not involve any target model or attack method. 
Hence, the meta generator only needs to be trained once, and it can be combined with different black-box attack methods to attack different black-box models without re-training, which corroborates the flexibility and generalization ability of the proposed framework.

\begin{table}
	\centering
	\caption{Comparison with Transfer-based methods on the ImageNet dataset.}
	\label{tab:transfer}
	\scriptsize
    \begin{tabular*}{\hsize}{@{}lccc@{\hspace{7pt}}c@{}}
	    \toprule
	    Target model $\rightarrow$ & ResNet-18 & VGG-16 & WRN-50 & Inception-V3 \\
		Attack Method $\downarrow$ & ASR & ASR & ASR & ASR \\
		\midrule
		PGD~\cite{PGD} & 35.5\% & 29.6\% & 36.6\% & 15.7\% \\
		PGD + MCG w/o surrogate & 56.8\% & 70.0\% & 49.5\% & \bf{26.4\%} \\
		PGD + MCG w/ fixed surrogate & \bf{57.9\%} & \bf{70.4\%} & \bf{50.1\%} & \bf{26.4\%} \\
		\midrule
		MI~\cite{MIFGSM} & 56.1\% & 62.0\% & 66.8\% & 26.4\% \\
		MI + MCG w/o surrogate & 62.6\% & 71.6\% & 67.6\% & 45.1\% \\
		MI + MCG w/ fixed surrogate & \bf{63.2\%} & \bf{73.6\%} & \bf{69.3\%} & \bf{45.4\%} \\
		\midrule
		TIMI~\cite{TIMIFGSM} & 56.7\% & \bf{63.4\%} & \bf{62.3\%} & 42.5\% \\
		TIMI + MCG w/o surrogate & 66.4\% & 62.1\% & 58.3\% & 42.7\% \\
		TIMI + MCG w/ fixed surrogate & \bf{66.7\%} & \bf{63.4\%} & 59.4\% & \bf{45.4\%} \\
		\midrule
		DI~\cite{DIFGSM} & 44.2\% & 42.0\% & 43.3\% & 20.1\% \\
		DI + MCG w/o surrogate & 68.5\% & \bf{80.0\%} & 67.1\% & 49.6\% \\
		DI + MCG w/ fixed surrogate & \bf{69.2\%} & \bf{80.0\%} & \bf{68.7\%} & \bf{50.7\%} \\
		\bottomrule
	\end{tabular*}
\end{table}

\vspace{1mm}
\noindent\textbf{Attacking Defended Models.}
To verify the effectiveness of the proposed method against adversarial defense models, we perform experiments of attacking various defended models trained with different defense strategies, including JPEG-Compression~\cite{guo2018countering}, random noise perturbation~\cite{SND}, and adversarial training~\cite{freeadv}.

JPEG-Compression defense (\ie, JPEG-Compress-WRN-50) attempts to remove the influence of adversarial examples through the compression process.
Small-Noise-Defense (\ie, SND-WRN-50) is specially designed for query-based attack via introducing additional random noise to hinder the attacker to estimate gradients correctly. 
As shown in Table~\ref{tab:imagenet_defense}, 
when attacking JPEG-Compress-WRN-50 and SND-WRN-50, the performance of NES, CG-Attack, SimBA-DCT, and MetaAttack drops sharply.
In contrast, the combination with our framework greatly improves their performance in all the three metrics.  
For example, when attacking JPEG-Compress-WRN-50 with CG-Attack, the ASR is only $39\%$ and the median query number is $1161$, which fails in most attacks. 
Our framework boosts its ASR to $91.7\%$ and reduces its median query number to $1$. 
Besides, our method also reduces the mean and median query numbers for Square and Signhunter. 

Adversarial training enlarges the difference of the classification boundaries between the robust model and the vanilla model and greatly limits the model-level adversarial transferability.
As shown in Table~\ref{tab:imagenet_defense}, when attacking the models of adversarial training, the attack performance of all methods drops a lot compared to attacking the vanilla models.  
Our method can still improve the attack performance in most cases, though the improvements are not as significant as those of attacking the vanilla models.

\begin{table*}[h]
	\centering
	
	\caption{Untargeted Attack trained on the CIFAR-10 dataset and tested on the CIFAR-100 dataset in the open-set scenario.}
	\label{tab:cifar100_openset}
	\scriptsize
	\begin{tabular*}{\hsize}{@{}@{\extracolsep{\fill}}lcccccccccccc@{}}
		\toprule
		 Target model $\rightarrow$ &\multicolumn{3}{c}{ResNet-18} & \multicolumn{3}{c}{VGG-16} & \multicolumn{3}{c}{WRN-50} & \multicolumn{3}{c}{Inception-V3} \\
		\cline{2-4}\cline{5-7}\cline{8-10}\cline{11-13}
		Attack Method $\downarrow$ & ASR & Mean & Median & ASR & Mean & Median & ASR & Mean & Median & ASR & Mean & Median \\
		\midrule
		
		NES~\cite{IlyasEAL18} & 96.1\% & 2054.0 & 1555.0 & 84.3\% & 1691.0 & 904.0 & 89.5\% & 2229.6 & 1534.0 & 94.6\% & 2044.7 & 1471.0 \\
MCG + NES & \bf97.0\% & \bf773.1 & \bf1.0 & \bf88.5\% & \bf910.5 & \bf1.0 & \bf94.0\% & \bf833.7 & \bf1.0 & \bf95.7\% & \bf1025.2 & \bf1.0 \\\midrule
             CG-Attack~\cite{feng2020boosting}& 99.4\% & 226.2 & \bf1.0 & \bf98.1\% & 294.7 & \bf1.0 & 98.4\% & 285.2 & \bf1.0 & 99.1\% & 330.8 & \bf1.0 \\
MCG + CG-Attack & \bf99.5\% & \bf135.1 & \bf1.0 & 97.8\% & \bf244.8 & \bf1.0 & \bf98.7\% & \bf163.5 & \bf1.0 & \bf99.6\% & \bf196.5 & \bf1.0 \\\midrule
SimBA-DCT~\cite{GuoGYWW19} & \bf99.2\% & 222.8 & 84.0 & 96.8\% & 321.4 & 66.0 & 97.4\% & 268.1 & 86.5 & 98.5\% & 231.6 & 76.0 \\
MCG + SimBA-DCT & \bf99.2\% & \bf119.5 & \bf1.0  & \bf97.0\% & \bf218.3 & \bf1.0 & \bf98.0\% & \bf155.2 & \bf1.0 & \bf98.9\% & \bf184.1 & \bf1.0 \\\midrule
             SignHunter~\cite{Al-DujailiO20} & \bf100.0\% & 88.2 & 28.0 & \bf99.6\% & 136.7 &  22.0 & \bf99.8\% & 134.5 & 28.0 & \bf100.0\%  & 117.8 & 32.0 \\ 
MCG + SignHunter & \bf100.0\% & \bf57.8 & \bf1.0 & \bf99.6\% &\bf 110.2 & \bf1.0 & \bf99.8\%  & \bf94.1 & \bf1.0 &\bf100.0\% & \bf92.0 & \bf1.0 \\ \midrule
Square~\cite{ACFH2020square} & \bf99.9\% & 103.8 & 17.0 & \bf99.5\% &213.9 & 24.0 & \bf99.8\% & 178.1 & 16.0 &99.8\%  &121.2 & 15.0 \\ 
MCG + Square & \bf99.9\% & \bf47.9 & \bf1.0 & \bf99.5\% & \bf92.7& \bf1.0 &\bf99.8\% &\bf83.1& \bf1.0 &\bf99.9\% & \bf75.6 & \bf1.0 \\\midrule
             MetaAttack~\cite{DuZZYF20} & \bf100.0\% & 518.9 & 443.0 &99.7\% &591.4 & 443.0 &99.9\% &610.1& 447.0&\bf100.0\% &523.6 & 445.0 \\
MCG + MetaAttack & 99.9\% &\bf227.4 & \bf1.0 &\bf99.9\% &\bf396.4& \bf1.0 &\bf100.0\% &\bf303.6& \bf1.0 &\bf100.0\% &\bf291.4& \bf1.0 \\ 
		\bottomrule
	\end{tabular*}
\end{table*}

\vspace{1mm}
\noindent\textbf{Comparison with Transfer-based Methods.}
To verify the effectiveness of example-level adversarial transferability boosted by meta learning, we perform an untargeted attack experiment to compare our meta generator with several transfer-based attack methods in the transfer attack  setting, \ie, no information from the black-box target model is used for fine-tuning.

Our meta generator contains a pre-training stage of the c-Glow  
and the pre-training is performed by using an attack method to generate adversarial examples for training. 
The quality of the adversarial examples affects the performance a lot. 
In other experiments, the attack method is PGD. 
Since the transfer-based methods can generate better transferable adversarial examples than PGD, to fairly compare with the transfer-based methods, here we pre-train the meta generator with using the adversarial examples generated by the three transfer-based methods, respectively. 

The results are shown in Table~\ref{tab:transfer}. 
`X + MCG w/o surrogate' means that we use `X' to pre-train the c-Glow and then directly use the meta generator to produce adversarial example without using the surrogate model. 
`X + MCG w/ fixed surrogate' means that after pre-training we use the surrogate model to fine-tune the meta generator first and then use the meta generator to produce adversarial examples. 
As shown in Table~\ref{tab:transfer}, directly using our meta generator has better performance than the transfer-based  methods in most cases.   
Fine-tuning the meta generator with the surrogate model can further improve the performance. 
Transfer-based methods may overfit the surrogate model due to that the adversarial example generation totally depends on the surrogate model. 
Differently, our meta generator captures the example-level transferability that can alleviate the overfitting issue. 
The generation is determined by both the learned prior of the meta generator and the surrogate model.

\vspace{1mm}
\noindent\textbf{Comparison with Combination-based Methods.}
As our method can be treated as a combination of transfer-based and query-based method, we compare with two combination-based black-box attack methods, \ie, TREMBA~\cite{Huang020} and AdvFlow~\cite{advflow}. 
Since they take advantage of model-level adversarial transferability to improve the performance and are often combined with evolution methods to adjust their distribution mapping.
For fairness, in the meta-test phase, we integrate the distribution adjustment method CG-Attack~\cite{feng2020boosting} into our framework for evaluation.

The results on ImageNet are shown in Table \ref{tab:imagenet_combination}.
Our method achieves the best performance under almost all attack cases in all the three metrics. 
Although other methods attempt to utilize the transferability between different models, they appear to be unstable in ASR when attacking different target models. 
For example, when attacking ResNet-18 and VGG-16 under targeted attack, the ASRs of TREMBA are $81.4\%$ and $91.1\%$. But when attacking WRN-50 and Inception-V3, the ASRs drop to $60.2\%$ and $13.3\%$. 
The results show that TREMBA cannot generalize well to attack different target models. 
Differently, our method can perform better in the generalization ability to attack different models.
The experimental results on CIFAR-10 are presented in the Appendix Sec.~B.1.

\subsection{Experiments in Open-set Attack Scenario}
In the closed-set attack scenario, the surrogate and target models share the same training dataset, \textit{i.e.,} the training dataset of the target model is visible to attackers, which is hard to achieve in the real attack scenario. 
In real-world scenarios, the surrogate model will share less common knowledge with the target model. This situation strongly increases the difficulty of attack. 
Therefore, in this section, we verify the effectiveness of our method in the open-set attack scenario where the surrogate and target models are trained on disjoint training datasets, and there is no overlap between the output categories of the two models. 
In this open-set attack setting, it is quite challenging for the attacker to transfer the limited prior to the unknown areas. 
Specifically, in our experiments, we employ two datasets and train the surrogate model on one dataset and the target model on another.
The training data of the meta generator is the same as the data used for the surrogate model.
The results of training on CIFAR-10 and testing on CIFAR-100 are shown in Table~\ref{tab:cifar100_openset}.
Please note that other experiments on ImageNet and OpenImage~\cite{openimage} are presented in the Appendix Sec.~B.2.

\begin{table}
    \centering
    \caption{Untargeted Attack against \emph{Imagga} tagging API.}
    \label{API}
    \vspace{-2mm}
    \begin{tabular*}{\hsize}{@{}@{\extracolsep{\fill}}lccccccc@{}}
    \toprule
    & \multicolumn{3}{c}{Baseline} & \multicolumn{3}{c}{Combined with MCG} \\
    \cline{2-4}\cline{5-7}
    Method & \scriptsize{ASR} & \scriptsize{Mean} & \scriptsize{Median} & \scriptsize{ASR} & \scriptsize{Mean} & \scriptsize{Median} \\
    \midrule
    NES~\cite{IlyasEAL18} & 44.0\% & 35.3 & 21.0 & \bf67.0\% & \bf8.8 & \bf1.0 \\
    CG-Attack~\cite{feng2020boosting} & 74.0\% & 33.5 & 21.0 & \bf81.0\% & \bf21.4 & \bf1.0 \\
    SimBA-DCT~\cite{GuoGYWW19} & 45.0\% & 93.8 & 56.0 & \bf57.0\% & \bf45.5 & \bf1.0 \\
    SignHunter~\cite{Al-DujailiO20} & 51.0\% & 45.3 & 20.5 & \bf82.0\% & \bf19.5 & \bf1.0 \\
    Square~\cite{ACFH2020square} & 49.0\% & 50.8 & 15.5 & \bf69.0\% & \bf13.7 & \bf1.0 \\
    MetaAttack~\cite{DuZZYF20} & 16.0\% & 230.1 & 95.0 & \bf61.0\% & \bf101.4 & \bf1.0 \\
    \bottomrule
    \end{tabular*}
\end{table}

Since the surrogate and target models are trained from different training sets with different categories, the effectiveness of model-level adversarial transferability is limited, which decreases the performance compared with the results in Table~\ref{tab:cifar10_closedset}.
Nevertheless, our framework still works in boosting the existing black-box attack methods in the open-set setting. 
MCG can reduce the query cost of attacks and improve the ASR in most cases, which demonstrates that the meta generator can be fast-adapted to attack different target models across different datasets.

\subsection{Experiments of Attacking Real-World API} 
To verify the effectiveness of our framework in real-world scenarios, we perform an experiment of attacking the Imagga Tagging API\footnote{https://imagga.com/solutions/auto-tagging}. 
The model of Imagga is trained on an unknown dataset of over $3,000$ types of daily-life objects. 
Given a query image, the API will return a list of possible labels as well as the corresponding confidence scores. 
We randomly select $100$ images from the validation set of ImageNet and set the query limit to $500$.
We define the goal of untargeted attack as removing the top-3 labels of the benign images. 
As the images are from ImageNet, the pre-trained surrogate model can be used to fine-tune the meta generator. 
$\mathcal{L}_{adv}$ is set as the maximal score of the top-3 labels.
The adapted generator is then used to generate an initial perturbation for existing black-box attack methods to attack the API.  
As shown in Table~\ref{API}, 
the performance of all methods is significantly improved through the combination with our framework. 
Specifically, the median query numbers are decreased to $1$s for all methods and the mean query numbers are also highly improved. 
These results demonstrate that our framework is applicable in real-world scenarios.

\begin{table}
	\centering
	
	\caption{
	Ablation study on the ImageNet dataset. All methods in the table are combined with Square attack for untargeted attack. 
	}
	\label{tab:ablation}
	\vspace{-2mm}
	\scriptsize
	\begin{tabular*}{\hsize}{@{}lc@{\hspace{8pt}}c@{\hspace{8pt}}cc@{\hspace{8pt}}c@{\hspace{8pt}}c@{}}
		\toprule
		Target Model $\rightarrow$ &\multicolumn{3}{c}{ResNet-18} & \multicolumn{3}{c}{VGG-16} \\
		\cline{2-4}\cline{5-7}
		Attack Method $\downarrow$ & \scriptsize{FASR} & \scriptsize{Mean} & \scriptsize{Median} & \scriptsize{FASR} & \scriptsize{Mean} & \scriptsize{Median} \\
		\midrule
		Flow & 35.2\% & 48.6 & 13.0 & 46.1\% & 39.1 & 4.0 \\
		MCG w/ fixed surrogate & 57.9\% & 35.3 & \bf1.0 & 70.4\% & 30.0 & \bf1.0 \\
		{MCG w/ fine-tuned surrogate} & \bf60.1\% & \bf31.7 & \bf1.0 & \bf71.3\% & \bf24.8 & \bf1.0 \\
		\bottomrule
	\end{tabular*}
\end{table}

\subsection{Ablation Study}

To verify the effectiveness of the meta training and the fine-tuning stages in the meta-test, we conduct experiments of untargeted attacks on ImageNet for ablation study. 
The results are shown in Table~\ref{tab:ablation}. 
`Flow' is the method that directly uses the perturbations to learn a conditional glow (c-Glow) model, rather than using the adversarial loss or the parameter update strategy in meta learning. 
The perturbations are generated by applying Projected Gradient Descent (PGD)~\cite{PGD} attack to the surrogate model.  
During testing, no fine-tuning is conducted.  
`MCG w/ fixed surrogate' means that during testing we fine-tune the meta generator with a fixed surrogate model.
`MCG w/ fine-tuned surrogate' means that during testing we update the surrogate model first by exploiting the query feedback from the target model, and then we use the updated surrogate model to fine-tune the meta generator. 
All these methods are combined with Square attack to query the target model. 
To evaluate the effectiveness of the initial perturbations, we use the first Attack Success Rate (FASR) as the metric instead of ASR. 
FASR means the success rate of straightforwardly using the perturbation generated by the generators to attack the target model. 

As shown in Table~\ref{tab:ablation}, compared to Flow, 
MCG w/ fixed surrogate achieves much better performance in all the three metrics. 
The FASRs and the median query numbers are significantly improved. 
The results demonstrate the effectiveness of the meta training formulation, which improves the example-level transferability by capturing more effective generic prior of how to attack different samples. The provided initial perturbation is better than that from Flow.  
Moreover, compared to MCG w/ fixed surrogate, MCG w/ fine-tuned surrogate gains further improvements in the FASR and the attack efficiency, which corroborates the effectiveness of the historical attack information, \textit{i.e.,} we can get a better-adapted generator by transferring the information of the target model to the surrogate model.

\section{Conclusion}
We propose a novel framework for black-box attack by formulating it as a meta-learning problem to improve the example-level adversarial transferability as well as the efficiency of attack.  
As the architecture and parameters of the black-box target model are unknown, we propose to perform the meta training with a surrogate by leveraging the model-level adversarial transferability. 
Since the standard meta-test process cannot be applied to the black-box attack, we propose a three-stage attack pipeline to fine-tune the meta model, including fine-tuning the surrogate model with historical attack information of the target model, fine-tuning the meta generator for the benign image with the updated surrogate model, and serving as the initialization to boost off-the-shelf black-box attack methods.   
Comprehensive experiments, including the closed-set and open-set scenarios as well as attacking online APIs, demonstrate the effectiveness of the proposed model.


\bibliographystyle{IEEEtran}
\bibliography{egbib}

\begin{thebibliography}{10}
\providecommand{\url}[1]{#1}
\csname url@samestyle\endcsname
\providecommand{\newblock}{\relax}
\providecommand{\bibinfo}[2]{#2}
\providecommand{\BIBentrySTDinterwordspacing}{\spaceskip=0pt\relax}
\providecommand{\BIBentryALTinterwordstretchfactor}{4}
\providecommand{\BIBentryALTinterwordspacing}{\spaceskip=\fontdimen2\font plus
\BIBentryALTinterwordstretchfactor\fontdimen3\font minus
  \fontdimen4\font\relax}
\providecommand{\BIBforeignlanguage}[2]{{%
\expandafter\ifx\csname l@#1\endcsname\relax
\typeout{** WARNING: IEEEtran.bst: No hyphenation pattern has been}%
\typeout{** loaded for the language `#1'. Using the pattern for}%
\typeout{** the default language instead.}%
\else
\language=\csname l@#1\endcsname
\fi
#2}}
\providecommand{\BIBdecl}{\relax}
\BIBdecl

\bibitem{GoodfellowSS14}
I.~J. Goodfellow, J.~Shlens, and C.~Szegedy, ``Explaining and harnessing
  adversarial examples,'' in \emph{Proc. Int. Conf. Learn. Represent.}, 2015.

\bibitem{moosavi2017universal}
S.-M. Moosavi-Dezfooli, A.~Fawzi, O.~Fawzi, and P.~Frossard, ``Universal
  adversarial perturbations,'' in \emph{Proc. IEEE Conf. Comput. Vis. Pattern
  Recog.}, 2017, pp. 1765--1773.

\bibitem{kurakin2018adversarial}
A.~Kurakin, I.~J. Goodfellow, and S.~Bengio, ``Adversarial examples in the
  physical world,'' in \emph{Artificial intelligence safety and
  security}.\hskip 1em plus 0.5em minus 0.4em\relax Chapman and Hall/CRC, 2018,
  pp. 99--112.

\bibitem{carlini2017cwattack}
N.~Carlini and D.~Wagner, ``Towards evaluating the robustness of neural
  networks,'' in \emph{2017 ieee symposium on security and privacy (sp)}.\hskip
  1em plus 0.5em minus 0.4em\relax IEEE, 2017, pp. 39--57.

\bibitem{laidlaw2019functional}
C.~Laidlaw and S.~Feizi, ``Functional adversarial attacks,'' in \emph{Proc.
  Adv. Neural Inform. Process. Syst.}, vol.~32, 2019.

\bibitem{duan2020adversarial}
R.~Duan, X.~Ma, Y.~Wang, J.~Bailey, A.~K. Qin, and Y.~Yang, ``Adversarial
  camouflage: Hiding physical-world attacks with natural styles,'' in
  \emph{Proc. IEEE Conf. Comput. Vis. Pattern Recog.}, 2020, pp. 1000--1008.

\bibitem{feng2021meta_pyhsical_attack}
W.~Feng, B.~Wu, T.~Zhang, Y.~Zhang, and Y.~Zhang, ``Meta-attack: Class-agnostic
  and model-agnostic physical adversarial attack,'' in \emph{Proc. IEEE Conf.
  Comput. Vis. Pattern Recog.}, 2021, pp. 7787--7796.

\bibitem{athalye2018synthesizing}
A.~Athalye, L.~Engstrom, A.~Ilyas, and K.~Kwok, ``Synthesizing robust
  adversarial examples,'' in \emph{Proc. Int. Conf. Mach. Learn.}, 2018, pp.
  284--293.

\bibitem{jan2019connecting}
S.~T. Jan, J.~Messou, Y.-C. Lin, J.-B. Huang, and G.~Wang, ``Connecting the
  digital and physical world: Improving the robustness of adversarial
  attacks,'' in \emph{Proc. of the AAAI Conf. on Artif. Intell.}, vol.~33,
  no.~01, 2019, pp. 962--969.

\bibitem{GuoGYWW19}
C.~Guo, J.~R. Gardner, Y.~You, A.~G. Wilson, and K.~Q. Weinberger, ``Simple
  black-box adversarial attacks,'' in \emph{Proc. Int. Conf. Mach. Learn.},
  2019, pp. 2484--2493.

\bibitem{Al-DujailiO20}
A.~Al{-}Dujaili and U.~O'Reilly, ``Sign bits are all you need for black-box
  attacks,'' in \emph{Proc. Int. Conf. Learn. Represent.}, 2020.

\bibitem{IlyasEAL18}
A.~Ilyas, L.~Engstrom, A.~Athalye, and J.~Lin, ``Black-box adversarial attacks
  with limited queries and information,'' in \emph{Proc. Int. Conf. Mach.
  Learn.}, 2018, pp. 2137--2146.

\bibitem{IlyasEM19}
A.~Ilyas, L.~Engstrom, and A.~Madry, ``Prior convictions: Black-box adversarial
  attacks with bandits and priors,'' in \emph{Proc. Int. Conf. Learn.
  Represent.}, 2019.

\bibitem{chen2017zoo}
P.-Y. Chen, H.~Zhang, Y.~Sharma, J.~Yi, and C.-J. Hsieh, ``Zoo: Zeroth order
  optimization based black-box attacks to deep neural networks without training
  substitute models,'' in \emph{Proceedings of the 10th ACM workshop on
  artificial intelligence and security}, 2017, pp. 15--26.

\bibitem{tu2019autozoom}
C.-C. Tu, P.~Ting, P.-Y. Chen, S.~Liu, H.~Zhang, J.~Yi, C.-J. Hsieh, and S.-M.
  Cheng, ``Autozoom: Autoencoder-based zeroth order optimization method for
  attacking black-box neural networks,'' in \emph{Proc. of the AAAI Conf. on
  Artif. Intell.}, vol.~33, no.~01, 2019, pp. 742--749.

\bibitem{liu2018signsgd}
S.~Liu, P.-Y. Chen, X.~Chen, and M.~Hong, ``signsgd via zeroth-order oracle,''
  in \emph{Proc. Int. Conf. Learn. Represent.}, 2018.

\bibitem{ACFH2020square}
M.~Andriushchenko, F.~Croce, N.~Flammarion, and M.~Hein, ``Square attack: a
  query-efficient black-box adversarial attack via random search,'' in
  \emph{Proc. Eur. Conf. Comput. Vis.}, 2020, pp. 484--501.

\bibitem{ChengDPSZ19}
S.~Cheng, Y.~Dong, T.~Pang, H.~Su, and J.~Zhu, ``Improving black-box
  adversarial attacks with a transfer-based prior,'' in \emph{Proc. Adv. Neural
  Inform. Process. Syst.}, vol.~32, 2019.

\bibitem{GuoYZ19}
Y.~Guo, Z.~Yan, and C.~Zhang, ``Subspace attack: Exploiting promising subspaces
  for query-efficient black-box attacks,'' in \emph{Proc. Adv. Neural Inform.
  Process. Syst.}, 2019, pp. 3820--3829.

\bibitem{DuZZYF20}
J.~Du, H.~Zhang, J.~T. Zhou, Y.~Yang, and J.~Feng, ``Query-efficient meta
  attack to deep neural networks,'' in \emph{Proc. Int. Conf. Learn.
  Represent.}, 2020.

\bibitem{LiLWZG19}
Y.~Li, L.~Li, L.~Wang, T.~Zhang, and B.~Gong, ``{NATTACK:} learning the
  distributions of adversarial examples for an improved black-box attack on
  deep neural networks,'' in \emph{Proc. Int. Conf. Mach. Learn.}, 2019, pp.
  3866--3876.

\bibitem{advflow}
H.~Mohaghegh~Dolatabadi, S.~Erfani, and C.~Leckie, ``Advflow: Inconspicuous
  black-box adversarial attacks using normalizing flows,'' in \emph{Proc. Adv.
  Neural Inform. Process. Syst.}, 2020, pp. 15\,871--15\,884.

\bibitem{Huang020}
Z.~Huang and T.~Zhang, ``Black-box adversarial attack with transferable
  model-based embedding,'' in \emph{Proc. Int. Conf. Learn. Represent.}, 2020.

\bibitem{brendel2018decision}
W.~Brendel, J.~Rauber, and M.~Bethge, ``Decision-based adversarial attacks:
  Reliable attacks against black-box machine learning models,'' in \emph{Proc.
  Int. Conf. Learn. Represent.}, 2018.

\bibitem{dong2019efficient}
Y.~Dong, H.~Su, B.~Wu, Z.~Li, W.~Liu, T.~Zhang, and J.~Zhu, ``Efficient
  decision-based black-box adversarial attacks on face recognition,'' in
  \emph{Proc. IEEE Conf. Comput. Vis. Pattern Recog.}, 2019, pp. 7714--7722.

\bibitem{rahmati2020geoda}
A.~Rahmati, S.-M. Moosavi-Dezfooli, P.~Frossard, and H.~Dai, ``Geoda: a
  geometric framework for black-box adversarial attacks,'' in \emph{Proc. IEEE
  Conf. Comput. Vis. Pattern Recog.}, 2020, pp. 8446--8455.

\bibitem{chen2020boosting}
W.~Chen, Z.~Zhang, X.~Hu, and B.~Wu, ``Boosting decision-based black-box
  adversarial attacks with random sign flip,'' in \emph{Proc. Eur. Conf.
  Comput. Vis.}, 2020, pp. 276--293.

\bibitem{chen2020rays}
J.~Chen and Q.~Gu, ``Rays: A ray searching method for hard-label adversarial
  attack,'' in \emph{Proc. Knowledge Discovery and Data Mining}, 2020, pp.
  1739--1747.

\bibitem{LiuMF19}
Y.~Liu, S.~Moosavi{-}Dezfooli, and P.~Frossard, ``A geometry-inspired
  decision-based attack,'' in \emph{Proc. Int. Conf. Comput. Vis.}, 2019, pp.
  4889--4897.

\bibitem{chen2020hopskipjumpattack}
J.~Chen, M.~I. Jordan, and M.~J. Wainwright, ``Hopskipjumpattack: A
  query-efficient decision-based attack,'' in \emph{IEEE Symposium on Security
  and Privacy}.\hskip 1em plus 0.5em minus 0.4em\relax IEEE, 2020, pp.
  1277--1294.

\bibitem{abs-2005-14137}
H.~Li, X.~Xu, X.~Zhang, S.~Yang, and B.~Li, ``{QEBA:} query-efficient
  boundary-based blackbox attack,'' in \emph{Proc. IEEE Conf. Comput. Vis.
  Pattern Recog.}, 2020, pp. 1218--1227.

\bibitem{ChengLCZYH19}
M.~Cheng, T.~Le, P.~Chen, H.~Zhang, J.~Yi, and C.~Hsieh, ``Query-efficient
  hard-label black-box attack: An optimization-based approach,'' in \emph{Proc.
  Int. Conf. Learn. Represent.}, 2019.

\bibitem{ChengSCC0H20}
M.~Cheng, S.~Singh, P.~H. Chen, P.~Chen, S.~Liu, and C.~Hsieh, ``Sign-opt: {A}
  query-efficient hard-label adversarial attack,'' in \emph{Proc. Int. Conf.
  Learn. Represent.}, 2020.

\bibitem{MIFGSM}
Y.~Dong, F.~Liao, T.~Pang, H.~Su, J.~Zhu, X.~Hu, and J.~Li, ``Boosting
  adversarial attacks with momentum,'' in \emph{Proc. IEEE Conf. Comput. Vis.
  Pattern Recog.}, 2018, pp. 9185--9193.

\bibitem{lin2019nesterov}
J.~Lin, C.~Song, K.~He, L.~Wang, and J.~E. Hopcroft, ``Nesterov accelerated
  gradient and scale invariance for adversarial attacks,'' in \emph{Proc. Int.
  Conf. Learn. Represent.}, 2020.

\bibitem{PapernotMG16}
N.~Papernot, P.~D. McDaniel, and I.~J. Goodfellow, ``Transferability in machine
  learning: from phenomena to black-box attacks using adversarial samples,''
  \emph{arXiv preprint arXiv:1605.07277}, 2016.

\bibitem{LiuCLS17}
Y.~Liu, X.~Chen, C.~Liu, and D.~Song, ``Delving into transferable adversarial
  examples and black-box attacks,'' in \emph{Proc. Int. Conf. Learn.
  Represent.}, 2017.

\bibitem{inkawhich2020perturbing}
N.~Inkawhich, K.~J. Liang, B.~Wang, M.~Inkawhich, L.~Carin, and Y.~Chen,
  ``Perturbing across the feature hierarchy to improve standard and strict
  blackbox attack transferability,'' in \emph{Proc. Adv. Neural Inform.
  Process. Syst.}, 2020, pp. 20\,791--20\,801.

\bibitem{Wu0X0M20}
D.~Wu, Y.~Wang, S.~Xia, J.~Bailey, and X.~Ma, ``Skip connections matter: On the
  transferability of adversarial examples generated with resnets,'' in
  \emph{{Proc. Int. Conf. Learn. Represent.}}, 2020.

\bibitem{tramer2017space}
F.~Tram{\`e}r, N.~Papernot, I.~Goodfellow, D.~Boneh, and P.~McDaniel, ``The
  space of transferable adversarial examples,'' \emph{arXiv preprint
  arXiv:1704.03453}, 2017.

\bibitem{yangcharacterizing}
Z.~Yang, L.~Li, X.~Xu, S.~Zuo, Q.~Chen, B.~Rubinstein, C.~Zhang, and B.~Li,
  ``Characterizing adversarial transferability via gradient orthogonality and
  smoothness,'' in \emph{Proc. Int. Conf. Mach. Learn. Worksh.}, 2020.

\bibitem{demontis2019adversarial}
A.~Demontis, M.~Melis, M.~Pintor, M.~Jagielski, B.~Biggio, A.~Oprea,
  C.~Nita-Rotaru, and F.~Roli, ``Why do adversarial attacks transfer?
  explaining transferability of evasion and poisoning attacks,'' in \emph{28th
  USENIX security symposium}, 2019, pp. 321--338.

\bibitem{wang2020unified}
X.~Wang, J.~Ren, S.~Lin, X.~Zhu, Y.~Wang, and Q.~Zhang, ``A unified approach to
  interpreting and boosting adversarial transferability,'' in \emph{Proc. Int.
  Conf. Learn. Represent.}, 2021.

\bibitem{wang2021enhancing}
X.~Wang and K.~He, ``Enhancing the transferability of adversarial attacks
  through variance tuning,'' in \emph{Proc. IEEE Conf. Comput. Vis. Pattern
  Recog.}, 2021, pp. 1924--1933.

\bibitem{huang2019enhancing}
Q.~Huang, I.~Katsman, H.~He, Z.~Gu, S.~Belongie, and S.-N. Lim, ``Enhancing
  adversarial example transferability with an intermediate level attack,'' in
  \emph{Proc. Int. Conf. Comput. Vis.}, 2019, pp. 4732--4741.

\bibitem{inkawhich2019transferable}
N.~Inkawhich, K.~Liang, L.~Carin, and Y.~Chen, ``Transferable perturbations of
  deep feature distributions,'' in \emph{Proc. Int. Conf. Learn. Represent.},
  2019.

\bibitem{qin2021meta_surrogate_model}
Y.~Qin, Y.~Xiong, J.~Yi, and C.-J. Hsieh, ``Training meta-surrogate model for
  transferable adversarial attack,'' \emph{arXiv preprint arXiv:2109.01983},
  2021.

\bibitem{yuan2021meta_gradient}
Z.~Yuan, J.~Zhang, Y.~Jia, C.~Tan, T.~Xue, and S.~Shan, ``Meta gradient
  adversarial attack,'' in \emph{Proc. Int. Conf. Comput. Vis.}, 2021, pp.
  7728--7737.

\bibitem{cGlow}
Y.~Lu and B.~Huang, ``Structured output learning with conditional generative
  flows,'' in \emph{Proc. of the AAAI Conf. on Artif. Intell.}, 2020, pp.
  5005--5012.

\bibitem{Tabak2010Density}
E.~Tabak and E.~Vanden-Eijnden, ``Density estimation by dual ascent of the
  log-likelihood,'' \emph{Communications in Mathematical Sciences}, vol.~8,
  no.~1, pp. 217--233, 2010.

\bibitem{Reptile}
A.~Nichol, J.~Achiam, and J.~Schulman, ``On first-order meta-learning
  algorithms,'' \emph{arXiv preprint arXiv:1803.02999}, 2018.

\bibitem{Adam}
D.~P. Kingma and J.~Ba, ``Adam: A method for stochastic optimization,'' in
  \emph{Proc. Int. Conf. Learn. Represent.}, 2015.

\bibitem{feng2020boosting}
Y.~Feng, B.~Wu, Y.~Fan, L.~Liu, Z.~Li, and S.~Xia, ``Boosting black-box attack
  with partially transferred conditional adversarial distribution,'' in
  \emph{Proc. IEEE Conf. Comput. Vis. Pattern Recog.}, 2022, pp.
  15\,095--15\,104.

\bibitem{krizhevsky2009}
A.~Krizhevsky, ``Learning multiple layers of features from tiny images,'' 2009.

\bibitem{RussakovskyDSKS15}
O.~Russakovsky, J.~Deng, H.~Su, J.~Krause, S.~Satheesh, S.~Ma, Z.~Huang,
  A.~Karpathy, A.~Khosla, M.~S. Bernstein, A.~C. Berg, and F.~Li, ``Imagenet
  large scale visual recognition challenge,'' \emph{Int. J. Comput. Vis.},
  2015.

\bibitem{HeZRS16}
K.~He, X.~Zhang, S.~Ren, and J.~Sun, ``Identity mappings in deep residual
  networks,'' in \emph{Proc. Eur. Conf. Comput. Vis.}, 2016, pp. 630--645.

\bibitem{HuangLMW17}
G.~Huang, Z.~Liu, L.~van~der Maaten, and K.~Q. Weinberger, ``Densely connected
  convolutional networks,'' in \emph{Proc. IEEE Conf. Comput. Vis. Pattern
  Recog.}, 2017, pp. 2261--2269.

\bibitem{SimonyanZ14a}
K.~Simonyan and A.~Zisserman, ``Very deep convolutional networks for
  large-scale image recognition,'' in \emph{Proc. Int. Conf. Learn.
  Represent.}, 2015.

\bibitem{HanKK17}
D.~Han, J.~Kim, and J.~Kim, ``Deep pyramidal residual networks,'' in
  \emph{Proc. IEEE Conf. Comput. Vis. Pattern Recog.}, 2017, pp. 6307--6315.

\bibitem{ZagoruykoK16}
S.~Zagoruyko and N.~Komodakis, ``Wide residual networks,'' in \emph{Brit. Mach.
  Vis. Conf.}, 2016.

\bibitem{szegedy2016rethinking}
C.~Szegedy, V.~Vanhoucke, S.~Ioffe, J.~Shlens, and Z.~Wojna, ``Rethinking the
  inception architecture for computer vision,'' in \emph{Proc. IEEE Conf.
  Comput. Vis. Pattern Recog.}, 2016, pp. 2818--2826.

\bibitem{guo2018countering}
C.~Guo, M.~Rana, M.~Cisse, and L.~van~der Maaten, ``Countering adversarial
  images using input transformations,'' in \emph{Proc. Int. Conf. Learn.
  Represent.}, 2018.

\bibitem{SND}
J.~Byun, H.~Go, and C.~Kim, ``Small input noise is enough to defend against
  query-based black-box attacks,'' \emph{Proc. Int. Conf. Learn. Represent.},
  2021.

\bibitem{freeadv}
A.~Shafahi, M.~Najibi, M.~A. Ghiasi, Z.~Xu, J.~Dickerson, C.~Studer, L.~S.
  Davis, G.~Taylor, and T.~Goldstein, ``Adversarial training for free!'' in
  \emph{Proc. Adv. Neural Inform. Process. Syst.}, 2019, pp. 3353--3364.

\bibitem{fastadv}
E.~Wong, L.~Rice, and J.~Z. Kolter, ``Fast is better than free: Revisiting
  adversarial training,'' in \emph{Proc. Int. Conf. Learn. Represent.}, 2019.

\bibitem{PGD}
A.~Madry, A.~Makelov, L.~Schmidt, D.~Tsipras, and A.~Vladu, ``Towards deep
  learning models resistant to adversarial attacks,'' in \emph{Proc. Int. Conf.
  Learn. Represent.}, 2018.

\bibitem{TIMIFGSM}
Y.~Dong, T.~Pang, H.~Su, and J.~Zhu, ``Evading defenses to transferable
  adversarial examples by translation-invariant attacks,'' in \emph{Proc. IEEE
  Conf. Comput. Vis. Pattern Recog.}, 2019, pp. 4312--4321.

\bibitem{DIFGSM}
C.~Xie, Z.~Zhang, Y.~Zhou, S.~Bai, J.~Wang, Z.~Ren, and A.~L. Yuille,
  ``Improving transferability of adversarial examples with input diversity,''
  in \emph{Proc. IEEE Conf. Comput. Vis. Pattern Recog.}, 2019, pp. 2730--2739.

\bibitem{openimage}
A.~Kuznetsova, H.~Rom, N.~Alldrin, J.~Uijlings, I.~Krasin, J.~Pont-Tuset,
  S.~Kamali, S.~Popov, M.~Malloci, A.~Kolesnikov \emph{et~al.}, ``The open
  images dataset v4,'' \emph{Int. J. Comput. Vis.}, pp. 1956--1981, 2020.

\bibitem{xiao2018generating}
C.~Xiao, B.~Li, J.-Y. Zhu, W.~He, M.~Liu, and D.~Song, ``Generating adversarial
  examples with adversarial networks,'' in \emph{Proc. Int. Joint Conf. on
  Artif. Intell.}, 2018, pp. 3905--3911.

\bibitem{jandial2019advgan++}
S.~Jandial, P.~Mangla, S.~Varshney, and V.~Balasubramanian, ``Advgan++:
  Harnessing latent layers for adversary generation,'' in \emph{Proc. Int.
  Conf. Comput. Vis. Worksh.}, 2019, pp. 2045--2048.

\bibitem{xiong2020improved}
Y.~Xiong and C.-J. Hsieh, ``Improved adversarial training via learned
  optimizer,'' in \emph{Proc. Eur. Conf. Comput. Vis.}, 2020, pp. 85--100.

\bibitem{SGM}
D.~Wu, Y.~Wang, S.-T. Xia, J.~Bailey, and X.~Ma, ``Skip connections matter: On
  the transferability of adversarial examples generated with resnets,'' in
  \emph{Proc. Int. Conf. Learn. Represent.}, 2020.

\end{thebibliography}

\vspace{-10mm}
\begin{IEEEbiography}
[{\includegraphics[width=0.8in,height=1in,clip,keepaspectratio]{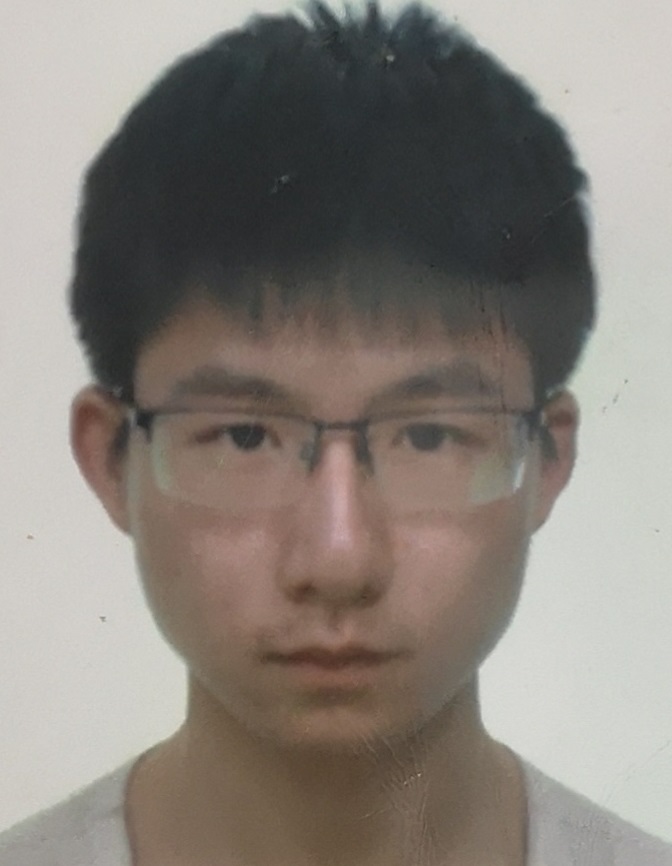}}]{Fei Yin} is currently a master student in Tsinghua Shenzhen International Graduate School, Tsinghua University. His current research interests include multimedia and computer vision.
\end{IEEEbiography}

\vspace{-20mm}
\begin{IEEEbiography}
[{\includegraphics[width=0.8in,height=1in,clip,keepaspectratio]{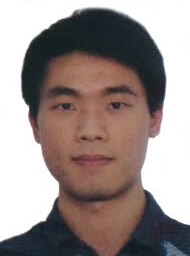}}]
{Yong Zhang}	
received the Ph.D. degree 	in pattern recognition and intelligent systems from the Institute of Automation, Chinese Academy of Sciences in 2018. From 2015 to 2017, he was a Visiting Scholar with the Rensselaer Polytechnic Institute. He is currently with the Tencent AI Lab. His research interests include computer vision and machine learning.
\end{IEEEbiography}

\vspace{-20mm}
\begin{IEEEbiography}
[{\includegraphics[width=0.8in,height=1in,clip,keepaspectratio]{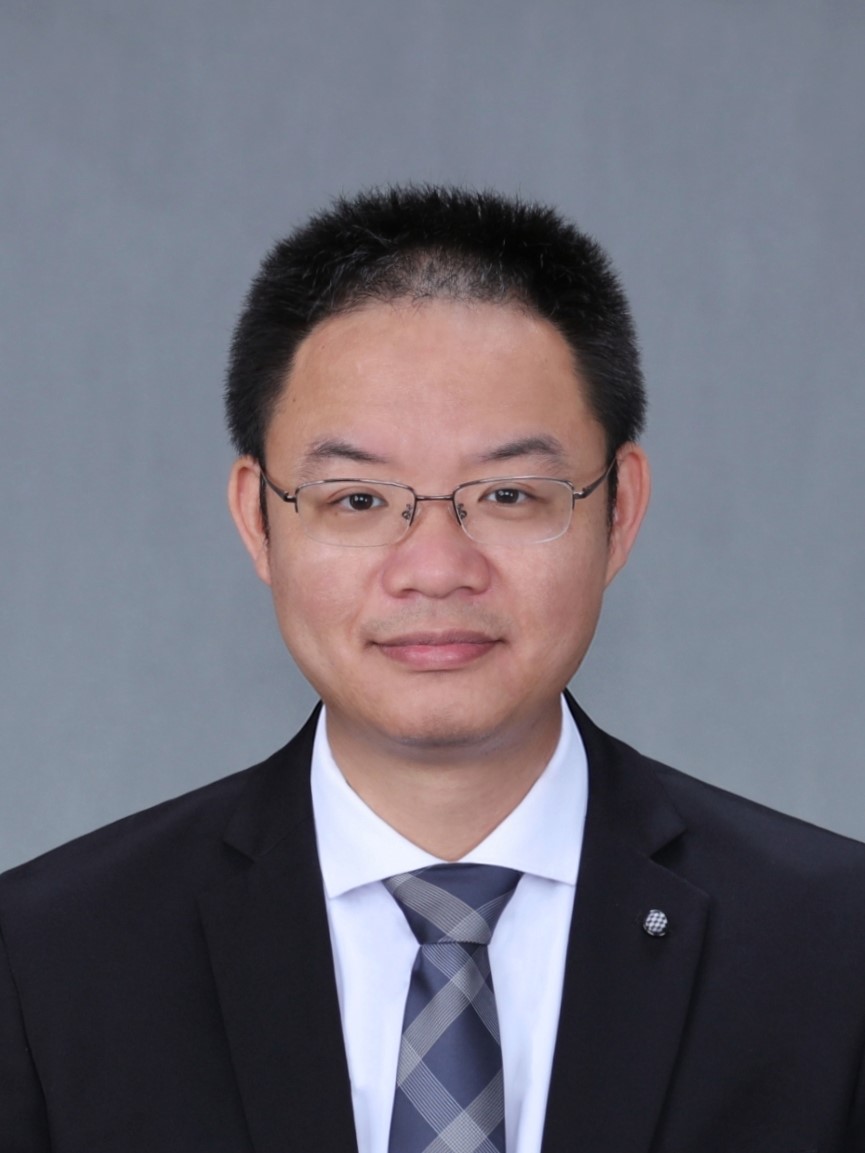}}]
{Baoyuan Wu} is an Associate Professor of School of Data Science, the Chinese University of Hong Kong, Shenzhen (CUHK-Shenzhen). He is also the director of the Secure Computing Lab of Big Data, Shenzhen Research Institute of Big Data (SBRID). He received the PhD degree from the National Laboratory of Pattern Recognition, Institute of Automation, Chinese Academy of Sciences, on June 2014. From November 2016 to August 2020, he was a Senior and Principal Researcher at Tencent AI lab. His research interests are AI security and privacy, machine learning, computer vision and optimization. 
\end{IEEEbiography}

\vspace{-12mm}
\begin{IEEEbiography}
[{\includegraphics[width=0.8in,height=1in,clip,keepaspectratio]{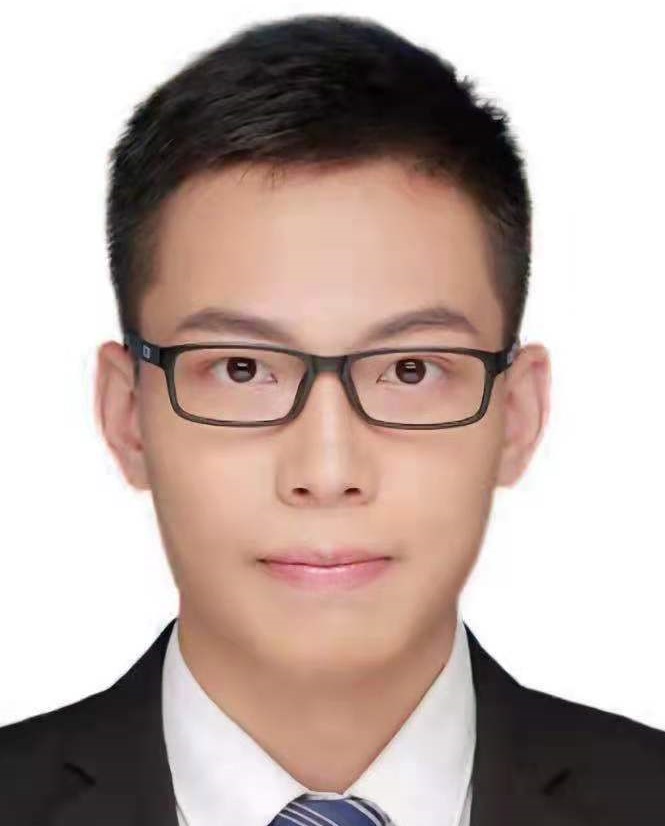}}]{Yan Feng} is currently a master student in Tsinghua Shenzhen International Graduate School, Tsinghua University. His current research interests include computer vision and AI security.
\end{IEEEbiography}

\vspace{-15mm}
\begin{IEEEbiography}[{\includegraphics[width=0.8in,height=1in,clip,keepaspectratio]{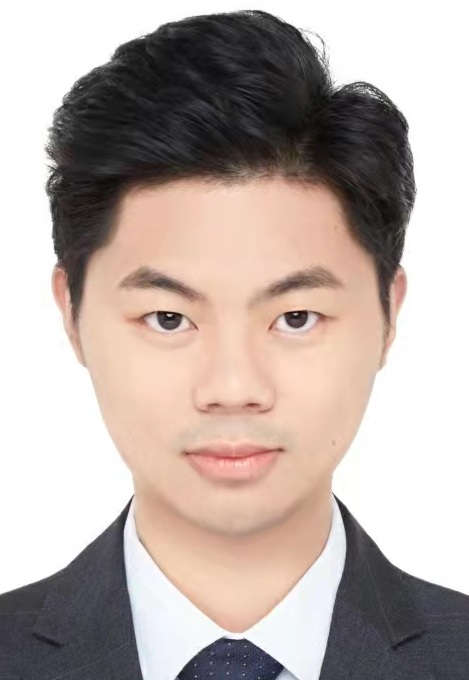}}]{Jingyi Zhang} is currently a master student in School of Computer Science and Engineering, University of Electronic Science and Technology of China.
His current research interests include multimedia and computer
vision.
\end{IEEEbiography}

\vspace{-15mm}
\begin{IEEEbiography}
	[{\includegraphics[width=0.8in,height=1in,clip,keepaspectratio]{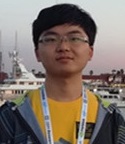}}]
	{Yanbo Fan}
	is currently a Senior Researcher at Tencent AI Lab. He received his Ph.D. degree from Institute of Automation, Chinese Academy of Sciences (CASIA), Beijing, China, in 2018, and his B.S. degree in Computer Science and Technology from Hunan University in 2013. His research interests are computer vision and machine learning.
\end{IEEEbiography}

\vspace{-15mm}
\begin{IEEEbiography}
[{\includegraphics[width=0.8in,height=1in,clip,keepaspectratio]{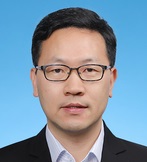}}]{Yujiu Yang} Yujiu Yang (Member, IEEE) received the Ph.D. degree from the Institute of Automation, Chinese Academy of Sciences. He is an Associate Professor with the Tsinghua Shenzhen International Graduate School, Tsinghua University. His research interests include natural language processing and computer vision.
\end{IEEEbiography}

\newpage
\appendices

\section{Method Analysis}

\subsection{Comparison with Methods Directly Using the Surrogate Model}

\noindent 
Our framework can combine the learned prior with different types of off-the-shelf query-based black-box attack methods in the meta-test phase and significantly boost their performance in terms of attack efficiency as well ASR. 

\begin{figure}[!htbp]
\begin{center}
\centerline{\includegraphics[width=1\linewidth]{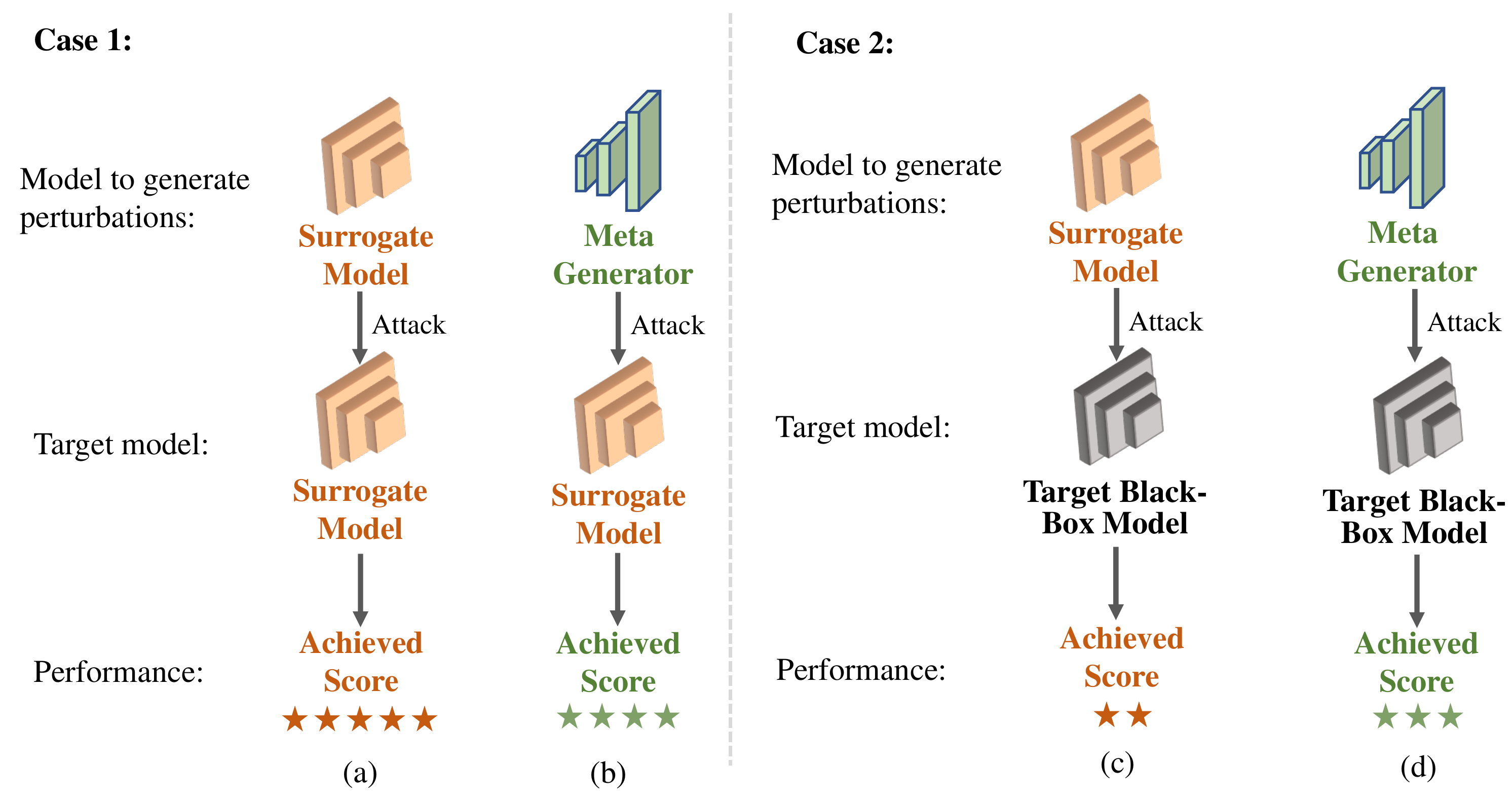}}
\vspace{-4mm}
\caption{Two cases for transfer attack.}
\label{fig:compare}
\end{center}
\vspace{-4mm}
\end{figure}

For a fair comparison with the transfer-based methods directly using the surrogate model, we only use the trained generator for evaluation.
Figure~\ref{fig:compare} presents the comparison between the surrogate model and our meta generator in two cases in the scenario of transfer attack, \textit{i.e.}, no query of the unknown target model.
In Case 1, the surrogate model is treated as the target model. 
Perturbations are generated by the surrogate model and our meta generator, respectively.
In this case, using the surrogate model to generate perturbation achieves much better performance than our meta generator (\textbf{\textit{e.g.,} ASR: PGD-100 100\% v.s. Ours 74\%. Surrogate model: ResNet-50, Target model: ResNet-50}). 
Because the unknown target model is the same as the surrogate model. 
Though our meta model is learned using the gradient from the surrogate, it does not try to learn mapping exactly from a sample to the gradient but learns the sample-dependent perturbation distribution by attacking a large set of samples, \textit{i.e.,} it captures some common properties among samples. 
Given a sample, our meta generator can provide a perturbation distribution that tells the probability of a sampled perturbation to be effective.  
It cannot predict the exact perturbation as the surrogate model. 
Therefore, in Case 1, the surrogate model wins. 

In Case 2, the unknown target model is different from the surrogate model. 
Given a sample, its generated perturbation is totally determined by the sample and the model itself. 
The perturbation seems to `overfit' the surrogate model as the gradient exactly comes from the surrogate model.  
Differently, our meta model generates the perturbation according to the sample and the learned prior (\textit{i.e.}, common properties among samples). 
Hence, the perturbation is generated by considering not one sample but a set of samples. 
It always generalizes better than the surrogate model in this case (\textbf{\textit{e.g.,} ASR: PGD-100 35.5\% v.s. Ours 56.8\%. Surrogate model: ResNet-50, Target model: ResNet-18}).

\begin{figure*}[!htbp]
\begin{center}
\centerline{\includegraphics[width=0.8\linewidth]{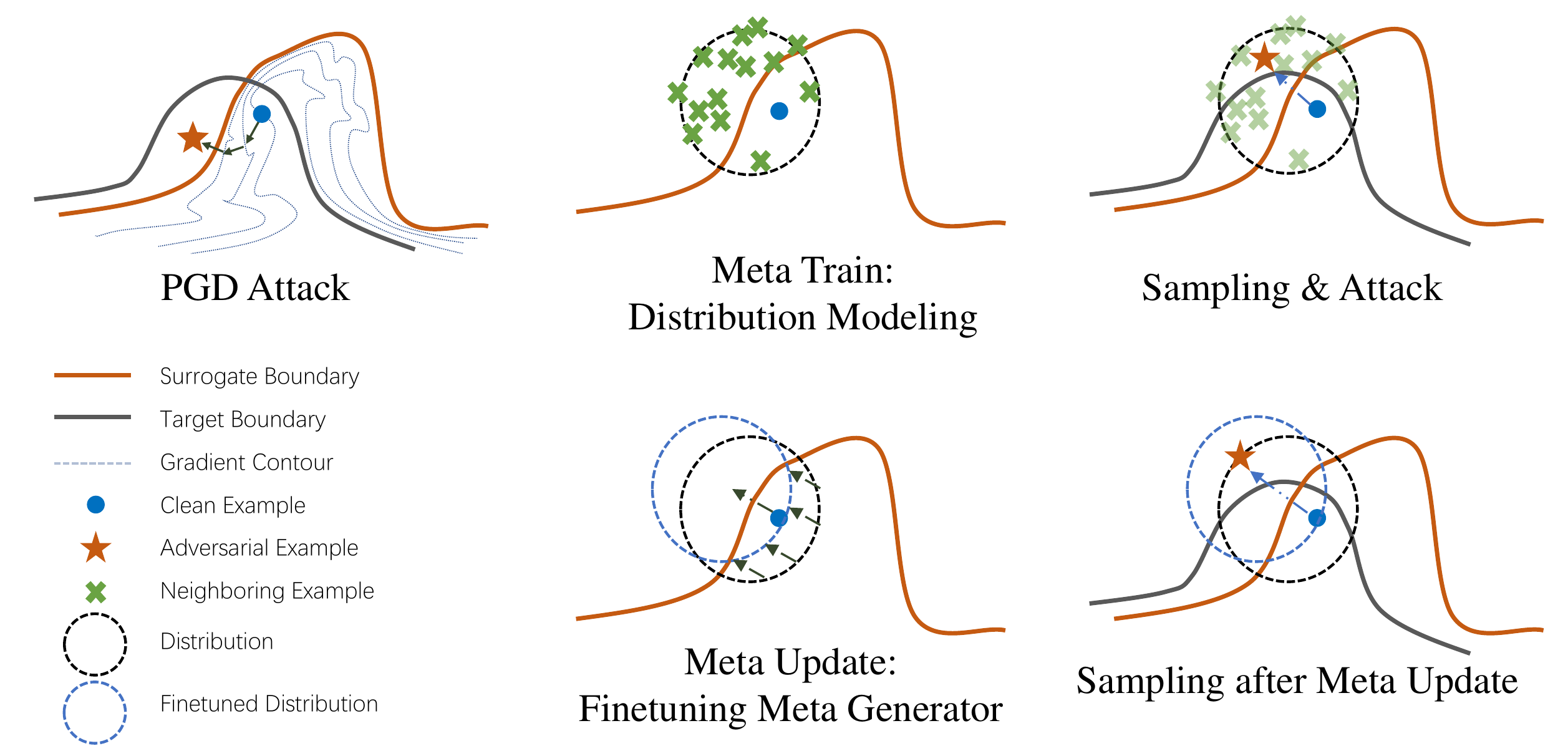}}
\vspace{-1mm}
\caption{Illustration of how the proposed method works.}
\label{fig:method}
\end{center}
\vspace{-4mm}
\end{figure*}

\subsection{Working Principle of MCG}

We provide an illustration of how our method works for better understanding in Figure~\ref{fig:method}. 
Figure~\ref{fig:method} \textcolor{black}{(a)} presents the attack by using the surrogate model via `\textbf{PGD}'.  The perturbation is generated according to the surrogate model and the clean example itself. 
Figure~\ref{fig:method} \textcolor{black}{(b)} shows that our meta generator captures the sample-dependent conditional distribution by performing a large set of attacking tasks involving a number of samples. 
The perturbation distribution is denoted by the black dash circle. 
Figure~\ref{fig:method} \textcolor{black}{(c)} shows that the meta generator is directly used to attack the unknown target model by sampling a perturbation with a high probability. It is a transfer attack. 
Figure~\ref{fig:method} \textcolor{black}{(d)} shows the meta update of the meta generator. 
The meta generator is combined with a query-based attack method. 
The query feedback is used to update the meta generator. 
Hence, the perturbation distribution is shifted from the black dash circle to the black dash circle. 
The black one is better as the update uses the feedback information about the target model. 
Figure~\ref{fig:method} \textcolor{black}{(e)} shows the usage of the updated meta generator. 
When combining with a query-based method that requires a perturbation as initialization, we can sample a perturbation with a high probability as the initialization.  
When combining with a query-based method that requires a distribution as initialization, then we can use the distribution represented by the generator as the initialization.  

According to Figure~\ref{fig:method}, our method learns the prior knowledge in the meta-train phase and updates the generator in the meta-test phase. 
It can be combined with off-the-shelf query-based methods. 
The meta update involves the query feedback that improves the meta generator. 
Our framework can boost the query-based methods in attack efficiency as well as ASR. 

\subsection{Advantages of Using c-Glow as Meta Generator}

\noindent 
c-Glow can model the exact log-likelihood of the underlying distribution, making it feasible to directly minimize the KL divergence between the approximated and real conditional adversarial distributions (CAD), rather than only optimizing the lower bound as in VAE models or finding an approximate maximum point in CAD as in learning-to-learn methods. 

In learning-to-learn methods, CNN and RNN generators are optimized based on the gradients of the classifiers, which means attackers can only pre-train these generators on surrogate classifiers, and then fully transfer their parameters for black-box attacks. As pointed out in \cite{feng2020boosting}, such a \textit{fully-transfer mechanism} will introduce the so-called \textit{surrogate bias} (due to differences in architectures and training datasets between surrogate and target models), which inevitably harms black-box attack performance. In contrast, as c-Glow consists of two parts of parameters, \ie, Gaussian and mapping parameters, we can utilize the \textit{partial transfer mechanism} to alleviate the surrogate bias. 

\subsection{Visualization of Adversarial Perturbations}

\begin{figure*}[!htbp]
\begin{center}
\centerline{\includegraphics[width=0.81\linewidth]{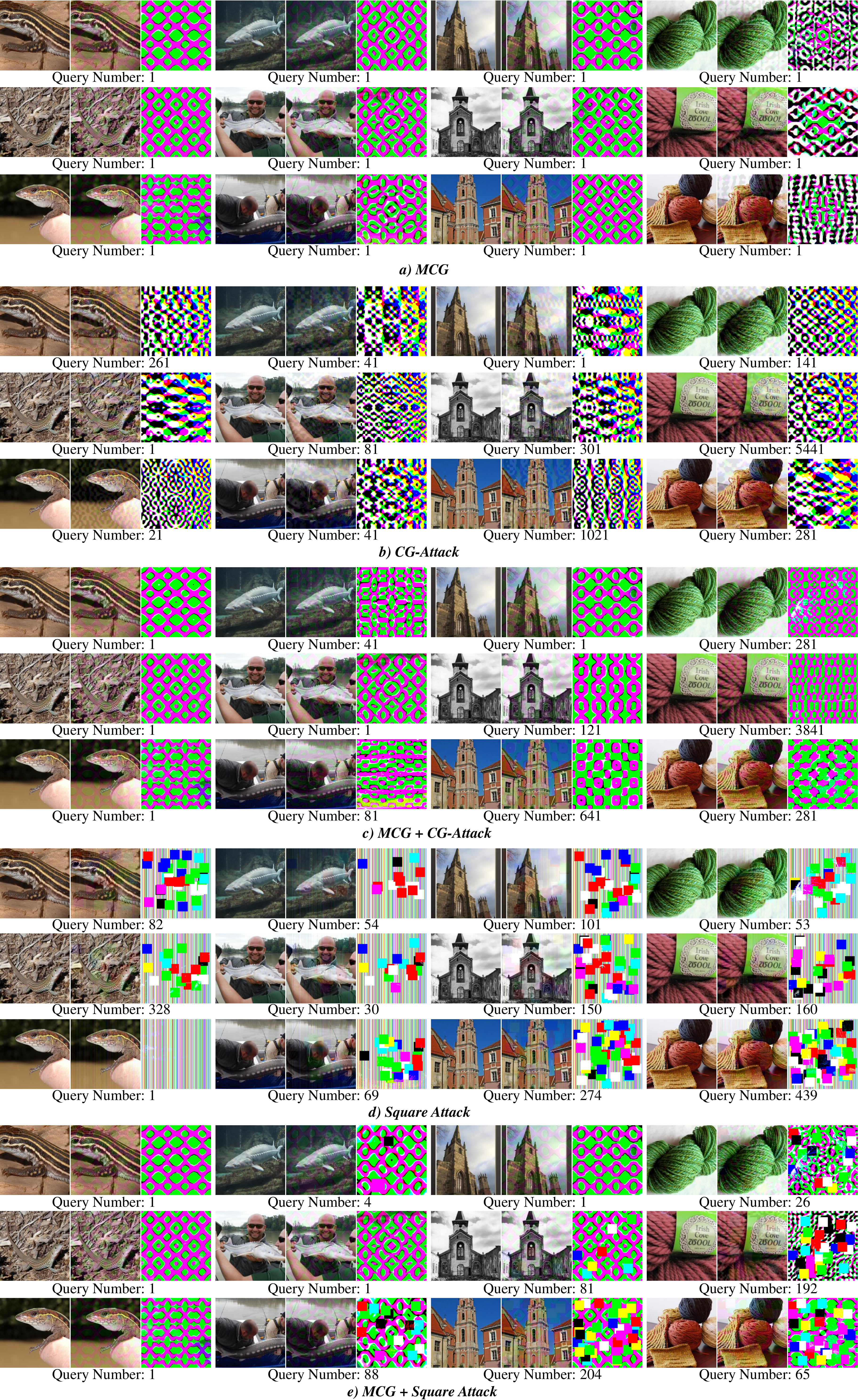}}
\caption{Visualization of the generated perturbations of the ImageNet dataset. 
In each triple block, each column represents the benign image, the adversarial example, and the perturbation, respectively.
The origin range of the perturbation is $[-0.05, 0.05]$ and we scale it to the range $[0, 255]$ for better visualization.
}
\label{fig:pattern}
\end{center}
\end{figure*}

\textcolor{rebuttal3}{
We present some visualization examples of five attack methods in Fig.~\ref{fig:pattern}. 
The experimental settings are as follows. 
We perform untargeted black-box attack with ResNet-50 as the surrogate model and ResNet-18 as the target model. 
Five attack methods are evaluated, including MCG (pure transfer), CG-Attack, MCG + CG-Attack, Square, and MCG + Square. 
The perturbation limit is set to $\ell_{\infty} \leq 0.05$. 
It is interesting to see that our proposed MCG is likely to generate nearly symmetric and rhombus-like patterns (see the top row in Fig. 1). Considering the extremely high transferability of these perturbations, they may provide good instances to analyze the characteristics of highly transferable perturbations.
However, we realize that these perturbations will vary across different clean images and different models. It requires more comprehensive evaluations and ingenious analysis tools/approaches to reveal some general characteristics. It will be explored in our future work.
}

\section{Additional Experiment Results}

\begin{table*}[t]
\centering
\caption{{Comparison with combination-based methods on the CIFAR-10 dataset}}
\label{tab:cifar10_combination}
\vspace{-2mm}
\scriptsize
\begin{tabular*}{\hsize}{@{}@{\extracolsep{\fill}}lcccccccccccc@{}}
    \toprule
    Target model $\rightarrow$ &\multicolumn{3}{c}{ResNet-PreAct-110} & \multicolumn{3}{c}{DenseNet-121} & \multicolumn{3}{c}{VGG-19} & \multicolumn{3}{c}{PyramidNet-110} \\
	\cline{2-4}\cline{5-7}\cline{8-10}\cline{11-13}
	Attack Method $\downarrow$ & ASR & Mean & Median & ASR & Mean & Median & ASR & Mean & Median & ASR & Mean & Median \\
	\midrule
	\multicolumn{13}{c}{\textit{\textbf{Untargeted Attack}}} \\
	\midrule
    TREMBA~\cite{Huang020} & 90.9\% & 120.7 & 64.0 & 97.8\% & 126.4 & 66.0 & 97.7\% & 125.5 & 63.0 & 97.9\% & 82.3 & 39.0 \\
    AdvFlow~\cite{advflow} & 97.2\% & 841.4 & 600.0 & \bf100.0\% & 1025.3 & 740.0 & 98.2\% & 1079.1 & 860.0 & 99.7\% & 857.5 & 560.0 \\
    MCG + CG-Attack & \textbf{100.0\%} & \textbf{41.8} & \textbf{1.0} & \textbf{100.0\%} & \textbf{21.3} & \textbf{1.0} & \textbf{100.0\%} & \textbf{38.8} & \textbf{1.0} & \textbf{100.0\%} & \textbf{12.8} & \textbf{1.0} \\    
    \midrule
    \multicolumn{13}{c}{\textit{\textbf{Targeted Attack}}} \\
    \midrule
    TREMBA~\cite{Huang020} & 91.2 & 1125.3  & 868.0 & 92.3 &  1123.4 & 879.0 & 96.5 & 1331.5 & 1142.0 & 98.1 & 1082.4 & 759.0 \\
    AdvFlow~\cite{advflow} & 98.6 & 911.7 & 820.0 & 96.3 & 1021.5 & 860.0 & \bf97.4 & 1144.1 & 940.0 & \bf100.0 & 908.1 & 820.0 \\
    MCG + CG-Attack & \textbf{100.0} & \textbf{430.8} & \textbf{181.0} & \textbf{99.8} & \textbf{608.5} & \textbf{241.0} & \textbf{97.4} & \textbf{944.1} & \textbf{381.0} & \textbf{100.0} & \textbf{290.6} & \textbf{141.0} \\
\bottomrule
\end{tabular*}
\end{table*}

\begin{table*}
	\centering
	
	\caption{{Untargeted attack on the ImageNet dataset in the open-set scenario.} 
	}
	\label{tab:imagenet_untarget_attack_openset_performance}
	\vspace{-2mm}
	\scriptsize
	\begin{tabular*}{\hsize}{@{}@{\extracolsep{\fill}}lcccccccccccc@{}}
		\toprule
		Target Model $\rightarrow$ &\multicolumn{3}{c}{ResNet-18} & \multicolumn{3}{c}{VGG-16} & \multicolumn{3}{c}{WRN-50} &
		\multicolumn{3}{c}{Inception-V3} \\
		\cline{2-4}\cline{5-7}\cline{8-10}\cline{11-13}
		Attack Method $\downarrow$ & ASR & Mean & Median & ASR & Mean & Median & ASR & Mean & Median & ASR & Mean & Median \\
		\midrule
		NES~\cite{IlyasEAL18} & 66.7\% & 4684.9 & 4516.0 & 68.5\% & 4903.7 & 4778.5 & 74.0\% & 4199.8 & 3760.0 & 94.9\% & 1989.0 & 1439.5 \\
		MCG + NES & \bf77.5\% & \bf4113.1 & \bf3845.0 & \bf77.4\% & \bf4392.6 & \bf4517.0 & \bf78.9\% & \bf3503.1 & \bf3089.0 & \bf96.8\% & \bf1773.5 & \bf1251.5 \\
		\midrule
		CG-Attack~\cite{feng2020boosting} & 50.2\% & 524.8 & 61.0 & 36.3\% & 476.3 & \bf41.0 & 55.1\% & 615.6 & 141.0 & 64.4\% & \bf497.3 & \bf41.0 \\
		MCG + CG-Attack & \bf57.7\% & \bf446.2 & \bf21.0 & \bf41.2\% & \bf468.6 & 61.0 & \bf55.2\% & \bf591.3 & \bf61.0 & \bf65.0\% & 521.4 & \bf41.0 \\
		\midrule
		SimBA-DCT~\cite{GuoGYWW19} & 20.9\% & 2123.4 & 2002.0 & 21.4\% & 2782.8 & 3202.0 & 33.3\% & 2947.2 & 3187.0 & 53.3\% & 2485.2 & 2283.0 \\
		MCG + SimBA-DCT & \bf42.9\% & \bf1182.9 & \bf829.0 & \bf35.2\% & \bf1006.6 & \bf669.0 & \bf52.4\% & \bf1293.4 & \bf1335.0 & \bf60.0\% & \bf1402.7 & \bf1343.0 \\
		\midrule
		Signhunter~\cite{Al-DujailiO20} & \bf100.0\% & \bf142.2 & \bf60.0 & 98.5\% & 391.6 & 145.0 & \bf100.0\% & 157.7 & 49.0 & \bf100.0\% & 129.5 & \bf45.0 \\
		MCG + Signhunter & \bf100.0\% & 161.0 & 64.0 & \bf99.2\% & \bf364.6 & \bf137.5 & \bf100.0\% & \bf153.0 & \bf44.0 & \bf100.0\% & \bf121.9 & 46.0 \\
		\midrule
		Square~\cite{ACFH2020square} & \bf100.0\% & 288.1 & 196.0 & \bf100.0\% & 632.6 & 336.0 & \bf100.0\% & 272.5 & 156.0 & \bf100.0\% & 207.1 & 126.0 \\
		MCG + Square & \bf100.0\% & \bf220.1 & \bf119.0 & \bf100.0\% & \bf572.3 & \bf271.0 & \bf100.0\% & \bf220.3 & \bf112.0 & \bf100.0\% & \bf175.8 & \bf97.0 \\
		\midrule
		MetaAttack~\cite{DuZZYF20} & 78.8\% & 5646.7 & 5951.0 & 75.7\% & 5485.2 & 5293.0 & 76.2\% & 5346.7 & 5505.5 & 93.6\% & 4168.6 & 4198.5 \\
		MCG + MetaAttack & \bf79.2\% & \bf4360.4 & \bf4416.0 & \bf76.6\% & \bf5078.9 & \bf5619.0 & \bf82.4\% & \bf4555.6 & \bf4412.0 & \bf94.5\% & \bf3896.9 & \bf3974.0 \\
		\bottomrule
	\end{tabular*}
\end{table*}

\begin{table*}[!htbp]
	\centering
	\caption{Untargeted attack evaluation on the OpenImage dataset.}
	\label{tab:openimage}
	\scriptsize
	\begin{tabular*}{\hsize}{@{}@{\extracolsep{\fill}}lcccccccccccc@{}}
		\toprule
		Target Model $\rightarrow$ &\multicolumn{3}{c}{ResNet-18} & \multicolumn{3}{c}{VGG-16} & \multicolumn{3}{c}{WRN-50} &
		\multicolumn{3}{c}{Inception-V3} \\
		\cline{2-4}\cline{5-7}\cline{8-10}\cline{11-13}
		Attack Method $\downarrow$ & ASR & Mean & Median & ASR & Mean & Median & ASR & Mean & Median & ASR & Mean & Median \\
		\midrule
		NES~\cite{IlyasEAL18} & 28.3\% & 4915.9 & 5041.0 & 40.4\% & 3894.7 & 3718.0 & 61.1\% & 4035.6 & 3823.0 & 86.9\% & 2671.2 & 1954.0 \\
        MCG + NES & \textbf{52.6\%} & \textbf{1306.7} & \textbf{281.0} & \textbf{69.9\%} & \textbf{764.3} & \textbf{1.0} & \textbf{74.0\%} & \textbf{1327.8} & \textbf{1.0} & \textbf{92.9\%} & \textbf{1712.8} & \textbf{23.0} \\
        \midrule
        CG-Attack~\cite{feng2020boosting} & 62.3\% & 2033.8 & 21.0 & 66.7\% & 2044.4 & 21.0 & 66.9\% & 1948.1 & \textbf{81.0} & 61.5\% & 2273.4 & 41.0 \\
        MCG + CG-Attack & \textbf{72.7\%} & \textbf{1337.4} & \textbf{1.0} & \textbf{82.5\%} & \textbf{1419.3} & \textbf{1.0} & \textbf{78.7\%} & \textbf{1201.2} & 101.0 & \textbf{75.5\%} & \textbf{1039.1} & \textbf{1.0} \\
        \midrule
        SimBA-DCT~\cite{GuoGYWW19} & 90.4\% & 374.201 & 17 & \textbf{94.9\%} & 254.7 & 10.0 & \textbf{92.5\%} & 301.4 & 12.0 & 83.2\% & 320.180 & 11.0 \\
        MCG + SimBA-DCT & \textbf{90.8\%} & \textbf{348.5} & \textbf{7.0} & 94.7\% & \textbf{226.2} & \textbf{1.0} & \textbf{92.5\%} & \textbf{250.7} & \textbf{1.0} & \textbf{83.9\%} & \textbf{249.6} & \textbf{1.0} \\
		\midrule
		Signhunter~\cite{Al-DujailiO20} & \textbf{99.7\%} & 331.8 & 11.0 & \textbf{100.0\%} & 149.2 & 7.0 & \textbf{100.0\%} & 233.3 & 13.0 & \textbf{100.0\%} & 259.9 & 8.0 \\
        MCG + Signhunter & 98.6\% & \textbf{267.9} & \textbf{6.0} & \textbf{100.0\%} & \textbf{100.9} & \textbf{1.0} & \textbf{100.0\%} & \textbf{159.9} & \textbf{1.0} & \textbf{100.0\%} & \textbf{247.7} & \textbf{4.0} \\
		\midrule
		Square~\cite{ACFH2020square} & \textbf{99.9\%} & 292.1 & 69.0 & \textbf{100.0\%} & 247.0 & 68.5 & \textbf{100.0\%} & 156.9 & 61.5 & \textbf{100.0\%} & 178.9 & 56.0 \\
        MCG + Square & 99.7\% & \textbf{230.7} & \textbf{24.0} & \textbf{100.0\%} & \textbf{180.9} & \textbf{1.0} & \textbf{100.0\%} & \textbf{93.9} & \textbf{1.0} & \textbf{100.0\%} & \textbf{123.9} & \textbf{8.5} \\
		\midrule
		MetaAttack~\cite{DuZZYF20} & 69.7\% & 4366.9 & 3964.0 & 79.9\% & 4029.4 & 3635.5 & 72.2\% & 4364.2 & 3754.0 & 81.4\% & 3689.8 & 3093.0 \\
        MCG + MetaAttack & \textbf{75.1\%} & \textbf{3033.8} & \textbf{1670.0} & \textbf{84.7\%} & \textbf{1637.9} & \textbf{1.0} & \textbf{76.9\%} & \textbf{1722.0} & \textbf{1.0} & \textbf{86.9\%} & \textbf{2351.1} & \textbf{41.0} \\
		\bottomrule
	\end{tabular*}
\end{table*}

\begin{table*}[t]
	\centering
	
	\caption{Untargeted Attack comparison with CNN and RNN-based generators on the CIFAR-10 dataset.}
	\label{tab:generator}
	\tiny
	\begin{tabular*}{\hsize}{@{}@{\extracolsep{\fill}}>{\color{rebuttal3}}l>{\color{rebuttal3}}c>{\color{rebuttal3}}c>{\color{rebuttal3}}c>{\color{rebuttal3}}c>{\color{rebuttal3}}c>{\color{rebuttal3}}c>{\color{rebuttal3}}c>{\color{rebuttal3}}c>{\color{rebuttal3}}c>{\color{rebuttal3}}c>{\color{rebuttal3}}c>{\color{rebuttal3}}c@{}}
	\toprule
	 Target model $\rightarrow$ &\multicolumn{3}{c}{\textcolor{rebuttal3}{ResNet-PreAct-110}} & \multicolumn{3}{c}{\textcolor{rebuttal3}{DenseNet-121}} & \multicolumn{3}{c}{\textcolor{rebuttal3}{VGG-19}} & \multicolumn{3}{c}{\textcolor{rebuttal3}{PyramidNet-110}} \\
	\cline{2-4}\cline{5-7}\cline{8-10}\cline{11-13}
	Attack Method $\downarrow$ & ASR & Mean & Median & ASR & Mean & Median & ASR & Mean & Median & ASR & Mean & Median \\
	\midrule
	Square & \bf100.0\% & 227.3 & 144.5 & \bf100.0\% & 260.3 & 159.0 & \bf100.0\% & 342.0 & 175.5 & \bf100.0\% & 165.5 & 100.5 \\
	CNN MCG + Square & \bf100.0\% & 67.4 & \bf1.0 & \bf100.0\% & 58.7 & \bf1.0 & \bf100.0\% & 120.8 & 24.0 & \bf100.0\% & 39.3 & \bf1.0 \\
	RNN MCG + Square & \bf100.0\% & 49.9 & \bf1.0 & \bf100.0\% & 69.4 & 6.0 & \bf100.0\% & 97.9 & 20.0 & \bf100.0\% & 33.1 & \bf1.0 \\
	c-Glow MCG + Square & \bf100.0\% & \bf39.7 & \bf1.0 & \bf100.0\% & \bf47.6 & \bf1.0 & \bf100.0\% & \bf57.9 & \bf1.0 & \bf100.0\% & \bf29.6 & \bf1.0 \\
	\bottomrule
	\end{tabular*}
\end{table*}

\begin{table*}[t]
	\centering
	
	\caption{Untargeted Attack comparison with MAML-based meta-learning strategy on the CIFAR-10 dataset.}
	\label{tab:maml}
	\tiny
	\begin{tabular*}{\hsize}{@{}@{\extracolsep{\fill}}>{\color{rebuttal3}}l>{\color{rebuttal3}}c>{\color{rebuttal3}}c>{\color{rebuttal3}}c>{\color{rebuttal3}}c>{\color{rebuttal3}}c>{\color{rebuttal3}}c>{\color{rebuttal3}}c>{\color{rebuttal3}}c>{\color{rebuttal3}}c>{\color{rebuttal3}}c>{\color{rebuttal3}}c>{\color{rebuttal3}}c@{}}
	\toprule
	 Target model $\rightarrow$ &\multicolumn{3}{c}{\textcolor{rebuttal3}{ResNet-PreAct-110}} & \multicolumn{3}{c}{\textcolor{rebuttal3}{DenseNet-121}} & \multicolumn{3}{c}{\textcolor{rebuttal3}{VGG-19}} & \multicolumn{3}{c}{\textcolor{rebuttal3}{PyramidNet-110}} \\
	\cline{2-4}\cline{5-7}\cline{8-10}\cline{11-13}
	Attack Method $\downarrow$ & ASR & Mean & Median & ASR & Mean & Median & ASR & Mean & Median & ASR & Mean & Median \\
	\midrule
	Square \cite{ACFH2020square} & \bf100.0\% & 227.3 & 144.5 & \bf100.0\% & 260.3 & 159.0 & \bf100.0\% & 342.0 & 175.5 & \bf100.0\% & 165.5 & 100.5 \\
	MCG-MAML + Square & \bf100.0\% & 42.2 & \bf1.0 & \bf100.0\% & \bf45.7 & \bf1.0 & \bf100.0\% & 107.1 & 18.0 & \bf100.0\% & \bf27.2 & \bf1.0 \\
	MCG-REPTILE + Square & \bf100.0\% & \bf39.7 & \bf1.0 & \bf100.0\% & 47.6 & \bf1.0 & \bf100.0\% & \bf57.9 & \bf1.0 & \bf100.0\% & 29.6 & \bf1.0 \\
	\bottomrule
	\end{tabular*}
\end{table*}

\begin{table*}[!htbp]
	\centering
	\caption{Validation of the extension with SGM. Untargeted attack evaluation on the ImageNet dataset.}
	\label{tab:sgm}
	\scriptsize
	\begin{tabular*}{\hsize}{@{}@{\extracolsep{\fill}}lcccccccccccc@{}}
		\toprule
		Target model $\rightarrow$ &\multicolumn{3}{c}{ResNet-18} & \multicolumn{3}{c}{VGG-16} & \multicolumn{3}{c}{WRN-50} & \multicolumn{3}{c}{Inception-V3} \\
		\cline{2-4}\cline{5-7}\cline{8-10}\cline{11-13}
		Attack Method $\downarrow$ & FASR & Mean & Median & FASR & Mean & Median & FASR & Mean & Median & FASR & Mean & Median \\
		\midrule
		MCG + Square & 60.1\% & 31.7 & \bf1.0 & 71.3\% & 24.8 & \bf1.0 & 51.3\% & 59.9 & \bf1.0 & 30.0\% & 123.8 & 24.0 \\
		SGM + MCG + Square & \bf67.9\% & \bf22.9 & \bf1.0 & \bf76.2\% & \bf22.6 & \bf1.0 & \bf65.5\% & \bf41.0 & \bf1.0 & \bf43.9\% & \bf88.2 & \bf6.0 \\
		\bottomrule
	\end{tabular*}
\end{table*}

\subsection{Comparison with Combination-based Methods on the CIFAR-10 dataset}

\vspace{1mm}
In Table~4 of the manuscript, we present the comparison with combination-based methods on the ImageNet dataset. 
Here, we provide the comparison results on the CIFAR-10 dataset. 
Competing methods are TREMBA~\cite{Huang020} and AdvFlow~\cite{advflow}. 
The results are shown in Table~\ref{tab:cifar10_combination}.
On CIFAR-10, our method achieves the best performance under all attack cases in all three metrics, which is consistent with the results on ImageNet.

\subsection{Additional Experiments in Open-set Attack Scenario}
In real-world scenarios, the surrogate model will share less common knowledge with the target model. 
This situation strongly increases the difficulty of attacking. 
In Sec.4.3 of the manuscript, we verify the adaptability of the proposed method across datasets via training on CIFAR-10 and testing on CIFAR-100.
Here, we provide additional experiments on ImageNet~\cite{RussakovskyDSKS15} and OpenImage~\cite{openimage}.

\vspace{1mm}
\noindent \textbf{Experiments on ImageNet.} 
\textcolor{black}{To simulate the open-set scenarios,} we first randomly select 10 classes from the $1,000$ classes of ImageNet and split the 10 classes into two groups evenly.  
We train the surrogate model on the training set of the one group of classes and train the target model on the training set of the other group of classes. 
The training data of the meta generator is the same as the data used for the surrogate model. 
The testing data is the validation set corresponding to the training categories of the target model.
The results of the untargeted attack on ImageNet are shown in Table~\ref{tab:imagenet_untarget_attack_openset_performance}.
The open-set attack on ImageNet is more challenging than that on CIFAR-10 as  
the median query numbers of several attack methods are relatively high, \textit{e.g.,} NES, SimBA-DCT, and MetaAttack. 
In this setting, our method can still improve their performance, especially the ASR for NES and SimBA-DCT. 

\vspace{1mm}
\noindent \textbf{Experiments of training on ImageNet and testing on OpenImage.}
To further verify the adaptability across different datasets, we perform another untargeted attack experiment by training the meta generator on the ImageNet dataset and testing it on the OpenImage dataset.
Target models (\textit{i.e.}, ResNet-18, VGG-16, WRN-50, and Inception-V3) are trained with the training data of 10 randomly selected classes of OpenImage.
The meta generator is trained with the training data of ImageNet guided by the surrogate model.
The results are shown in Table~\ref{tab:openimage}. 
Our MCG can reduce the query cost of attacks and improve the ASR in most cases, which demonstrates that the meta generator can be fast-adapted to different target models across different datasets.

\subsection{Comparison with CNN and RNN-based generators.}

\noindent \textcolor{rebuttal3}{
CNN and RNN-based generators can also be fine-tuned to mitigate the bias in our framework. 
In our framework the generator is used to capture the prior distribution of adversarial examples, which is a replaceable component. 
Other types of generators can be flexibly incorporated into the framework to replace the c-Glow. 
To compare the influence of different generators, we perform experiments by integrating the CNN or RNN-based generator into our framework to replace c-Glow.
}

\vspace{1mm}
\noindent \textcolor{rebuttal3}{
\textbf{Implementation details of the CNN-based generator.}
We follow \cite{xiao2018generating} and \cite{jandial2019advgan++} to re-implement the CNN-based generator. 
The backbone includes a feature extractor $f$, a generator network $G$, and a discriminator network $D$. 
We concatenate the feature $f(x)$ of image $x$ and a noise vector sampled from learnable mean parameters $z$. 
Then we feed the concatenation to the generator $G$. 
The generator $G$ predicts an adversary perturbation $x_{adv}$ corresponding to $x$. 
The discriminator $D$ distinguishes the output distribution of the generator with the real distribution.
The adversarial loss of the surrogate model (Eq.~4 of the manuscript) is also used. 
To bound the magnitude of perturbation, we minimize $L_{inf}$ bound norm of adversary perturbation.
The loss function is defined as:
\begin{equation}
    \mathcal{L} = \mathcal{L}_{\text{GAN}} + \alpha \mathcal{L}_{adv} + \beta \mathcal{L}_{inf},
\end{equation}
where
\begin{equation}
    \mathcal{L}_{GAN} = \mathbb{E}_{x}[\log \mathcal{D}(x) + \mathbb{E}_{x} \log (1 - \mathcal{D}(x + G(z, f(x)))],
\end{equation}
\begin{equation}
    \mathcal{L}_{adv} = \mathbb{E}_{x} [g_{w}(x + G(z, f(x)), t)],
\end{equation}
\begin{equation}
    \mathcal{L}_{inf} = \mathbb{E}_{x}\Vert G(z, f(x)) \Vert _{\infty}.
\end{equation}
$t$ is the target class and $g_{w}$ denotes the surrogate classifier.
}

\vspace{1mm}
\noindent \textcolor{rebuttal3}{
\textbf{Implementation details of the RNN-based generator.}
We follow \cite{xiong2020improved} to re-implement the RNN-based model. 
We flatten the input benign images into one-dimension feature vectors and directly feed the features into the RNN network $G$. 
The initial hidden state $h$ for the input sequence is sampled from the learnable mean parameters $z$. 
We reshape the output sequence from $G$ back to the corresponding spatial dimension and achieve the adversarial perturbation. 
Similarly, the adversarial loss $\mathcal{L}_{adv}$ and the bound loss $\mathcal{L}_{inf}$ are used to optimize the generator. 
The loss function is defined as:
\begin{equation}
    \mathcal{L} = \mathcal{L}_{adv} + \lambda \mathcal{L}_{inf}. 
\end{equation}
For the rest training and testing strategy, we keep the settings the same as in our manuscript. 
}

\vspace{1mm}
\noindent \textcolor{rebuttal3}{
\textbf{Experimental results.}
The experimental results are shown in Tab.~\ref{tab:generator}. 
We use Square~\cite{ACFH2020square} as the baseline method. 
It can be observed that both CNN-based and RNN-based generators can improve the performance of the baseline method considerably. 
The results demonstrate the scalablility of our framework, \textit{i.e.,} the c-Glow generator can be replaced by other types of generators. 
Comparing the three types of generator, Flow-based MCG achieves better performance than the other two. 
Moreover, overall the RNN-based generator performs slightly better than the CNN-based generator. 
}

\subsection{Comparison with MAML-based Meta-learning Methods.}

\noindent \textcolor{rebuttal3}{
Both MAML and REPTILE are powerful meta-learning algorithms aiming at optimizing for an initial representation that can be effectively fine-tuned.
MAML unrolls and differentiates through the computation graph of the gradient descent algorithm, while Reptile simply performs stochastic gradient descent  on each task, which makes Reptile take less computation and memory than MAML.
We perform an experiment to compare REPTILE-based MCG with MAML-based MCG in the untargeted attack scenario on the CIFAR-10 dataset. The baseline method is Square~\cite{ACFH2020square}. 
Results are shown in Tab.~\ref{tab:maml}. 
It can be observed that `MCG-MAML + Square' and `MCG-REPTILE + Square' achieve close performance. 
This comparison also demonstrates the flexibility of using different meta-learning algorithms in the proposed framework. 
}

\subsection{Extended Experiments with Skip Gradient Method}

Skip Gradient Method (SGM)~\cite{SGM} is a transfer-based attack method that excavates the internal gradient flow of skip-connection branches to generate more transferable perturbation.
Since we utilize the model-level adversarial transferability through the surrogate model, we can introduce the strategy of SGM into our framework to boost the attack performance. 
Similar to SGM, we change the backward gradient weights of skip-connection branches of our surrogate model to train the generator.
During the attacking process, we apply the same strategy to the surrogate model to fine-tune our meta generator.
We perform an experiment on ImageNet in the untargeted attack scenario with ResNet-50 as the surrogate model. 
The results are shown in Table~\ref{tab:sgm}. 
`MCG + Square' is our original method. `SGM + MCG + Square' means that we additional introduce the strategy of SGM. 
The results show the SGM strategy helps our model achieve improvements in both FASR and the query number.

\end{document}